\pdfoutput=1
\documentclass[11pt]{article}
\usepackage[]{naacl2021}
\usepackage{times}
\usepackage{latexsym}
\usepackage[T1]{fontenc}
\usepackage[utf8]{inputenc}
\usepackage{microtype}

\usepackage{url}
\usepackage{amsmath}
\usepackage{tabularx}
\usepackage{tikzsymbols}
\usepackage{xcolor}
\usepackage{graphicx}
\usepackage{enumitem}
\usepackage{color,colortbl}
\usepackage{soul}
\usepackage{amssymb}
\usepackage{pifont}
\usepackage{hyperref} 
\usepackage{cleveref} 
\usepackage{booktabs}
\usepackage{xcolor}
\usepackage{bbm}
\usepackage{subfig}
\usepackage[hang,flushmargin]{footmisc}

\usepackage{xcolor, soul}

\definecolor{mypink}{rgb}{0,0,0,1} % Uncomment for final version 
\newcommand{\kmy}[1]{\textcolor{black}{#1}}
\newcommand{\greencheckmark}[0]{\ding{52}}

\newcommand{\norm}[1]{\left\lVert#1\right\rVert}

\definecolor{Gray1}{rgb}{0.91,0.925, 0.937}
\definecolor{Gray2}{rgb}{0.87, 0.886, 0.902}
\definecolor{Gray3}{rgb}{0.808, 0.831, 0.855}
\definecolor{Gray4}{rgb}{0.678,0.71, 0.741}

\definecolor{Blue1}{rgb}{0.792, 0.941, 0.973}
\definecolor{Blue2}{rgb}{0.678, 0.91, 0.957}
\definecolor{Blue3}{rgb}{0.565, 0.878, 0.937}
\definecolor{Blue4}{rgb}{0.282, 0.749, 0.89}

\definecolor{Yellow1}{rgb}{1, 0.914, 0.306}
\definecolor{Yellow2}{rgb}{1, 0.886, 0.275}
\definecolor{Yellow3}{rgb}{1, 0.855, 0.239}

\definecolor{abhi}{cmyk}{0, 0, 0, 1} % Uncomment for final version
\newcommand{\abhi}[1]{\textcolor{abhi}{#1}}

\definecolor{InformationTheoretic}{HTML}{FED89F}
\definecolor{Geometric}{HTML}{98EAF8}
\definecolor{HigherOrder}{HTML}{F9BFDA}
\title{Domain Divergences: A Survey and Empirical Analysis}

\author{Abhinav Ramesh Kashyap\(^\dagger\), \ Devamanyu Hazarika\(^\dagger\), Min-Yen Kan\(^\dagger\), Roger Zimmermann \(^\dagger\) \\
\(^\dagger\)National University of Singapore, Singapore\\
\fontsize{10}{12}\texttt{\{abhinav, hazarika, kanmy, rogerz\}@comp.nus.edu.sg}\\ 
}

\date{}

\begin{document}
\maketitle
\begin{abstract}
Domain divergence plays a significant role in estimating the performance of a model in new domains.  While there is a significant literature on divergence measures, researchers find it hard to choose an appropriate divergence for a given NLP application.  We address this shortcoming by both surveying the literature and through an empirical study.  We develop a taxonomy of divergence measures consisting of three classes  --- Information-theoretic, Geometric, and Higher-order measures and identify the relationships between them. Further, to understand the common use-cases of these measures, we recognise three novel applications -- 1) Data Selection, 2) Learning Representation, and 3) Decisions in the Wild -- and use it to organise our literature.  From this, we identify that Information-theoretic measures are prevalent for 1) and 3), and Higher-order measures are more common for 2). To further help researchers choose appropriate measures to predict drop in performance -- an important aspect of  Decisions in the Wild, we perform correlation analysis spanning 130 domain adaptation scenarios, 3 varied NLP tasks and 12 divergence measures identified from our survey. To calculate these divergences, we consider the current contextual word representations (CWR) and contrast with the older distributed representations. We find that traditional measures over word distributions still serve as strong baselines, while higher-order measures with CWR are effective.
\end{abstract}

% **********************************
%           INTRODUCTION 
% **********************************
\section{Introduction}

% \textbf{General Context}
% \begin{itemize}
%     \item Machine learning model performance reduces in a new domain 
%     \item Performance in target domain depends on the divergence between domains 
% \end{itemize}

Standard machine learning models do not perform well when tested on data from a different target domain. The performance in a target domain largely depends on the domain divergence~\cite{ben2010theory} -- \abhi{a notion of distance between the two domains}. Thus, efficiently measuring and reducing divergence is crucial for adapting models to the new domain --- the topic of {\it domain adaptation}. Divergence also has practical applications in predicting the performance drop of a model when adapted to new domains~\cite{van-asch-daelemans-2010-using}, and in choosing among alternate models~\cite{xia2020predicting}. 

\abhi{Given its importance,} researchers have invested much effort to define and measure domain divergence. Linguists use register variation to capture varieties in text --  the difference between distributions of the prevalent features in two registers~\cite{biber_conrad_2009}. \abhi{Other measures include probabilistic measures like $\mathcal{H}$-divergence \cite{ben2010theory}, information theoretic measures like Jenssen-Shannon and Kullback-Leibler divergence~\cite{plank-van-noord-2011-effective,van-asch-daelemans-2010-using} and measures using higher-order moments of random variables like Maximum Mean Discrepancy (MMD) and Central Moment Discrepancy (CMD) \cite{Gretton2006, Zellinger2017CentralMD}.} The proliferation of divergence measures challenges researchers in choosing an appropriate measure for a given application.

% Abhinav: Moving the paragraph here seeing whether it makes sense according to Liangming's strong suggestions
% Abhinav: Repetitive from the previous paragraph
% Given a plethora of divergences, researchers are not clear about the right measure for a given application. 
To help guide best practices, we first comprehensively review the NLP literature on domain divergences. Unlike previous surveys, which focus on domain adaptation for specific tasks such as machine translation~\cite{chu-wang-2018-survey} and statistical (non-neural network) models \cite{Jiang2007ALS,Margolis2011ALR}, our work takes a different perspective. We study domain adaptation through the vehicle of {\it domain divergence measures}. First, we develop a taxonomy of divergence measures consisting of three groups: Information-Theoretic, Geometric, and Higher-Order measures. Further, to find the most common group used in NLP, we recognise three novel application areas of these divergences --- Data Selection, Learning Representations, and Decisions in the Wild and organise the literature under them.  We find that Information-Theoretic measures over word distributions are popular for Data Selection and Decisions in the wild, while Higher-order measures over continuous features are frequent for Learning representations.  

Domain divergence is a major predictor of performance in the target domain. A better domain divergence metric ideally predicts the corresponding performance drop of a model when applied to a target domain -- \abhi{a practical and important component of \textit{Decisions in the Wild}}. We further help researchers identify appropriate measures \abhi{for predicting performance drops}, through a correlation analysis over 130 domain adaptation scenarios and three standard NLP tasks: Part of Speech Tagging (POS), Named Entity Recognition (NER), and Sentiment Analysis and 12 divergence metrics from our literature review.  While information-theoretic measures over traditional word distributions are popular in the literature, are higher-order measures calculated over modern contextual word representations better indicators of performance drop? We indeed find that higher-order measures are superior, but traditional measures are still reliable indicators of performance drop. \abhi{The closest to our work is \cite{elsahar-galle-2019-annotate} who perform a correlation analysis. However, they do not compare against different divergence measures from the literature. Comparatively, we consider more tasks and divergence measures.}

In summary, our contributions are:
\vspace{-2.2mm}
\begin{itemize}[leftmargin=*]
\itemsep-0.37em 
    \item We review the literature from the perspective of domain divergences and their use-cases in NLP.
    \item We aid researchers to select appropriate divergence measure that indicate performance-drops, an important application of divergence measures.
\end{itemize}

% % **********************************
% %       DIVERGENCE AND TASKS
% % **********************************
%%%%%%%%%%%%%%%%%%%%%%%%%%%%%%%%%%%%%%
%       Taxonomy of metrics              %
%%%%%%%%%%%%%%%%%%%%%%%%%%%%%%%%%%%%%%
\section{A Taxonomy of Divergence Measures}

\begin{figure*}[t]
    \centering
    \includegraphics[width=\linewidth]{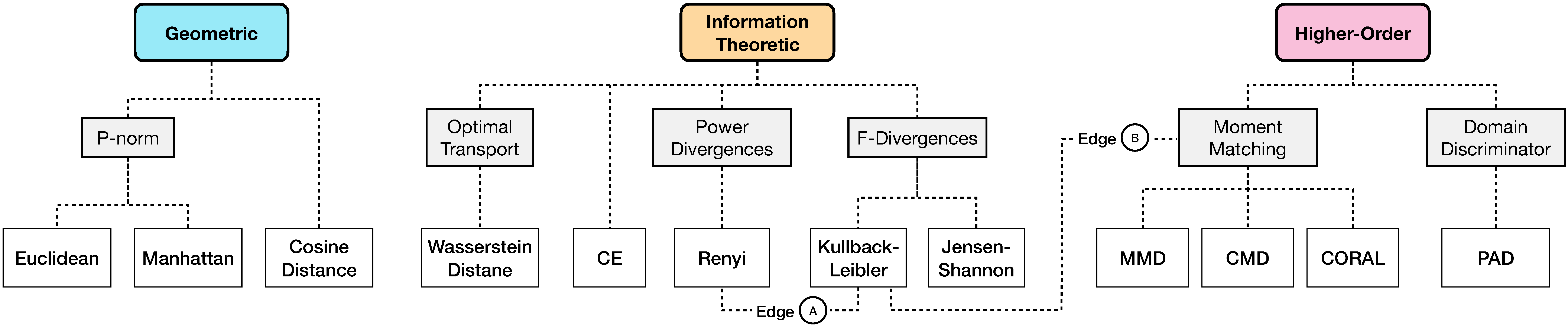}
    \caption{
    Taxonomy for divergence measures.  i) \textbf{Geometric} measures the distance between vectors in a metric space ii) \textbf{Information- theoretic} measures the distance between probability distributions and iii) \textbf{Higher-order} measures the distance between distributions considering higher moments or the distance between representations or their projections in a nonlinear space. Edge \raisebox{.5pt}{\textcircled{\raisebox{-.9pt} {A}}} indicates that Renyi divergence tends towards KL divergence as $\alpha \rightarrow 1$ and Edge \raisebox{.5pt}{\textcircled{\raisebox{-.9pt} {B}}} indicates KL-Div can be considered as matching first-order moment.
    }
    \label{fig:domain-adaptation-measures}
\end{figure*}

We devise a taxonomy for domain divergence measures, shown in \Cref{fig:domain-adaptation-measures}. It contains three main classes. Individual measures belong to a single class, where relationships can exist between measures from different classes.  We provide detailed description of individual measures in Appendix~\ref{sec:background}. 

\textbf{Geometric measures} calculate the distance between two vectors in a metric space. As a divergence measure, they calculate the distance between features  ($tf.idf$, continuous representations, etc.) extracted from instances of different domains. The P-norm is a generic form of the distance between two vectors, where Manhattan \textit{(p=1)} and Euclidean distance \textit{ (p=2)} are common. Cosine (Cos) uses the cosine of the angle between two vectors to measure similarity and 1-Cos measures distance.
% MinArXiV: do you need to back this claim up with argument or citation?
%Abhi: Added this
Geometric measures are easy to calculate, but are ineffective in a high dimensional space as all distances appear the same \cite{surprisingbehavior}.

{\bf Information-theoretic measures} captures the distance between probability distributions. For example, cross entropy over n-gram word distributions are extensively used in domain adaptation for machine translation. $f$-divergence \cite{csiszar1972} is a general family of divergences where $f$ is a convex function. Different formulations of the $f$ function lead to KL and JS divergence. \citet{chen-cardie-2018-multinomial} show that reducing $f$-divergence measure is equivalent to reducing the PAD measures (see next section).  Another special case of $f$-divergence is the family of $\alpha$ divergences, where KL-Div is a special case of $\alpha$ divergence. Renyi Divergence is a member of the $\alpha$-divergences and tends towards KL-Div as $\alpha \rightarrow 1$ (Edge \raisebox{.5pt}{\textcircled{\raisebox{-.9pt} {A}}});
Often applied to optimal transport problems, Wasserstein distance measures the amount of work needed to convert one probability distribution to the other as distance and is used extensively for domain adaptation. KL-Div is also related to Cross Entropy (CE). In this paper, CE refers to measures based on entropy.
% MinArXiV: be consistent with your capitalisation.  I see KL-Div, KL div, KL-div, Kullback-Leibler .  Pick a form and stick with it consistently.
% AbhiArXiV: Fixed it
% MinArXiV: same problem with tf.idf.
% AbhiArXiV: fixed it 

\textbf{Higher-Order} measures consider matching higher order moments of random variables or divergence in a projected space. Their properties are amenable to end-to-end learning based domain adaptation and recently have been extensively adopted. 
% MinArXiV: why and justify.  This is just finger pointing.  Or at least cite.
% AbhiArXiV:
Maximum Mean Discrepancy~(MMD) is one such measure which considers matching first order moments of variables in a Reproducible Kernel Hilbert Space. On the other hand, CORAL~\cite{DBLP:series/acvpr/SunFS17} considers second order moments and CMD~\cite{Zellinger2017CentralMD} considers higher order moments. CORAL and CMD are desirable because they avoid computationally expensive kernel matrix computations.  KL-Div can also be considered as matching the first-order moment~\cite{Zellinger2017CentralMD}; Edge \raisebox{.5pt}{\textcircled{\raisebox{-.9pt} {B}}}. Proxy-A-Distance (PAD) measures the distance between source and target distributions via the error of a classifier in target domain samples as source domain samples \cite{BenDavid2006}.

% MinArXiV: not really a category.  Not having much use is not a theoretical organization.  Push these into the three categories.  
% AbhiArXiV: Converting this into a closing paragraph. We have a column called Others in the table. There should be one paragraph to explain what that is

\noindent
A few other measures do not have ample support in the literature. These include information-theoretic measures such as Bhattacharya coefficient, higher-order measures like PAD* \cite{elsahar-galle-2019-annotate}, Word Vector Variance (WVV), and Term Vocabulary Overlap (TVO)~\cite{dai-etal-2019-using}. Our taxonomy synthesises the diversity and the prevalence of the divergence measures in NLP.

% MinArXiV: somehow this summary doesn't add anything.  Need it to be more tight and signal worthy to include.
% AbhiArXiV: Any suggestions on what can be changed and how to make it better ?
% There are a wide variety of divergence measures used in the literature. This domain divergence taxonomy compares these divergence measures based on their merits and limits. It further provides a unified view, by highlighting the similarities and relationship between these measures.

%%%%%%%%%%%%%%%%%%%%%%%%%%%%%%%%%%%%%%%%%%%%%%%%%%%%%%%%%%%%%%%%%%%%%%%
% Applications of Domain Divergence
%%%%%%%%%%%%%%%%%%%%%%%%%%%%%%%%%%%%%%%%%%%%%%%%%%%%%%%%%%%%%%%%%%%%%%%

\section{Applications of Divergence Measures}

Our key observation of the literature is that there are three primary families of applications of divergences (cf. \Cref{tab:divergence-table} in the appendix):
($i$) \textbf{Data Selection}: selects a subset of text from a source domain that shares similar characteristics as target domain. The selected subset is then used to learn a target domain model. ($ii$) \textbf{Learning Representations}: aligns source and target domain distributions and learn domain-invariant representations. ($iii$) \textbf{Decisions in the Wild}: helps practitioners predict the performance or drops in performance of a model in a new target domain.

\abhi{We limit the scope our survey to works that focus on divergence measures. We only consider unsupervised domain adaptation (UDA) --  where there is no annotated data available in the target domain. It is more practical yet more challenging.  For a complete treatment of neural networks and UDA in NLP, refer to \cite{Ramponi2020NeuralUD}.  Also, we do not treat multilingual work. While cross-lingual transfer can be regarded as an extreme form of domain adaptation, measuring the distance between languages requires different divergence measures, outside our purview.}

\begin{table*}[t!]
    \small % Sets the font size to small on the entire table
    \centering
    \begin{tabular}{p{3.7cm}|p{1.4cm}|ccccc|cc|ccc|c}
         \hline
         \textbf{Paper} & \textbf{Task(s)} & \multicolumn{5}{|c|}{\textbf{Information-Theoretic}} & \multicolumn{2}{|c|}{\textbf{ Geometric}} & \multicolumn{3}{|c|}{\textbf{ Higher-Order}} & \textbf{Others} \\
         \hline
          &  & {\tiny KL} & {\tiny JS} & {\tiny Renyi} & {\tiny CE} & {\tiny Wass.} & {\tiny Cos} & {\tiny P-Norm} & {\tiny PAD} & {\tiny CMD}  & {\tiny MMD} &  -\\ 
         \hline
         \rowcolor{Gray2}
         \multicolumn{13}{c}{\textbf{\textsc{Data Selection}}} \\
         \hline
         \rowcolor{Gray1}
         \cite{plank-van-noord-2011-effective} & Par, POS & \greencheckmark & \greencheckmark & \greencheckmark &  & & \greencheckmark & & & & &  \\ 
         \rowcolor{Gray2}
         \cite{dai-etal-2019-using} & NER & & & & & & & & & & & \greencheckmark \\ 
         \rowcolor{Gray2}
         \cite{ruder-plank-2017-learning} & SA, NER, Par  & & \greencheckmark & \greencheckmark & & & \greencheckmark & \greencheckmark & \greencheckmark & & &  \greencheckmark \\ 
         \rowcolor{Gray3}
         \cite{Ruder2017DataSS} & SA  & & \greencheckmark & & & & \greencheckmark & & \greencheckmark & & & \greencheckmark  \\ 
         \rowcolor{Gray3}
         \cite{DBLP:conf/icdm/Remus12} & SA  & & \greencheckmark & & & & & & & & &  \\ 
         \rowcolor{Gray4}
         \cite{lu-etal-2007-improving} & SMT & & & & & & \greencheckmark & & & & & \\ 
         \rowcolor{Gray4}
         \cite{zhao-etal-2004-language} & SMT & & & & & & \greencheckmark & & & & & \\ 
         \rowcolor{Gray4}
         \cite{yasuda-etal-2008-method} & SMT & & & & \greencheckmark & & & & & & & \\ 
         \rowcolor{Gray4}
         \cite{moore-lewis-2010-intelligent} & SMT & & & & \greencheckmark & & & & & & &  \\ 
         \rowcolor{Gray4}
         \cite{axelrod-etal-2011-domain} & SMT & & & & \greencheckmark & & & & & & &  \\ 
         \rowcolor{Gray4}
         \cite{duh-etal-2013-adaptation} & SMT & & & & \greencheckmark & & & & & & &  \\ 
         \rowcolor{Gray4}
         \cite{liu-etal-2014-effective} & SMT & & & & \greencheckmark & & & & & & &  \\ 
         \rowcolor{Gray4}
         \cite{Wees2017DynamicDS} & NMT & & & & \greencheckmark & & & & & & &  \\
         \rowcolor{Gray4}
         \cite{silva-etal-2018-extracting} & NMT  & & & & & & & & & & & \greencheckmark \\ 
         \rowcolor{Gray4}
         \cite{Aharoni2020UnsupervisedDC} & NMT  & & & & & & \greencheckmark & & & & &  \\ 
         \rowcolor{Gray4}
         \cite{wang-etal-2017-sentence} & NMT & & & & & & & \greencheckmark & & & &  \\ 
         \rowcolor{Gray4}
         \cite{carpuat-etal-2017-detecting} & NMT  & & & & & & & & & & & \greencheckmark \\ 
         \rowcolor{Gray4}
         \cite{vyas-etal-2018-identifying} & NMT  & & & & & & & & & & & \greencheckmark \\ 
         \rowcolor{Gray4}
         \cite{chen-huang-2016-semi} & SMT & & & & & & & & & & & \greencheckmark \\
         \rowcolor{Gray4}
         \cite{chen-etal-2017-cost} & NMT & & & & & & & & & & & \greencheckmark \\
         \hline 
         
        %%%%%%%%%%%%%%%%%%%% Learning Representations %%%%%%%%%%%%%%%%%%%%%%
        \rowcolor{Blue1}
         \multicolumn{13}{c}{\textbf{\textsc{learning representations}}}  \\
         \hline 
         \rowcolor{Blue1}
         \cite{Ganin2015DomainAdversarialTO} & SA & & & & & & & & \greencheckmark & & &  \\
         \rowcolor{Blue1}
         \cite{kim-etal-2017-adversarial} & Intent-clf & & & & & & & & \greencheckmark & & & \\
         \rowcolor{Blue1}
         \cite{liu-etal-2017-adversarial} & SA & & & & & & & & \greencheckmark & & & \\ 
         \rowcolor{Blue1}
         \cite{li-etal-2018-whats} & Lang-ID & & & & & & & & \greencheckmark & & & \\
         \rowcolor{Blue1}
         \cite{chen-cardie-2018-multinomial}& SA & & & & & & & & & \greencheckmark & &  \\ 
         \rowcolor{Blue1}
         \cite{Zellinger2017CentralMD}& SA & & & & & & & & & \greencheckmark & &  \\ 
         \rowcolor{Blue1}
         \cite{peng-etal-2018-cross}& SA & & & & & & & & & \greencheckmark & &  \\ 
         \rowcolor{Blue1}
         \cite{Wu2019DualAC} & SA & & & & & & & & \greencheckmark & & & \\
         \rowcolor{Blue1}
          \cite{dingmultidomain} & Intent-Clf  & & & & & & & & \greencheckmark & & & \\ 
          \rowcolor{Blue1}
          \cite{shah-etal-2018-adversarial} & Question sim & & & & & \greencheckmark & & & \greencheckmark & & & \\ 
          \rowcolor{Blue1}
          \cite{zhu-etal-2019-adversarial} & Emo-Regress & & & & & \greencheckmark & & & & & & \\ 
          \rowcolor{Blue2}
          \cite{gui-etal-2017-part} & POS & & & & & & & & \greencheckmark & & & \\ 
          \rowcolor{Blue2}
          \cite{zhou-etal-2019-dual} & NER & & & & & & &  & \greencheckmark & & & \\
         \rowcolor{Blue2}
         \cite{cao-etal-2018-adversarial} & NER & & & & & & & & \greencheckmark & & & \\ 
         \rowcolor{Blue2}
         \cite{wang-etal-2018-label} & NER & & & & & & & & & & \greencheckmark &  \\ 
         \rowcolor{Blue3}
         \cite{gu-etal-2019-improving} & NMT & & & & & & & & \greencheckmark & & & \\ 
         \rowcolor{Blue3}
         \cite{britz-etal-2017-effective} & NMT & & & & & & & &\greencheckmark & & & \\ 
         \rowcolor{Blue3}
         \cite{zeng-etal-2018-multi} & NMT & & & & & & & & \greencheckmark & & & \\ 
         \rowcolor{Blue3}
         \cite{Wang2019GoFT} & NMT & & & & & & & & \greencheckmark  & & & \\ 
         \hline
        %%%%%%%%%%%%%%%%%%%Applications in the Wild%%%%%%%%%%%%%%%%%%%%%    
         \rowcolor{Yellow1}
         \multicolumn{13}{c}{\textbf{\textsc{decisions in the wild}}} \\
         \hline 
         \rowcolor{Yellow1}
         \cite{ravi-etal-2008-automatic} & Parsing & & & & \greencheckmark & & & & & & & \\ 
         \rowcolor{Yellow3}
         \cite{elsahar-galle-2019-annotate} & SA, POS  & & & & & & & & \greencheckmark & & & \greencheckmark \\
         \rowcolor{Yellow3}
         \cite{ponomareva2012biographies} & SA & \greencheckmark & \greencheckmark & & & & \greencheckmark & & & & & \greencheckmark \\ 
         \rowcolor{Yellow3}
         \cite{van-asch-daelemans-2010-using} & POS & \greencheckmark & & \greencheckmark & & & & \greencheckmark & & & & \greencheckmark \\ 
         \hline 
    \end{tabular}
    \caption{Prior works using divergence measures for \textit{Data Selection}, \textit{Learning Representations} and \textit{Decisions in the Wild}. Tasks can be \textit{Par}: dependency parsing, \textit{POS}: Parts of Speech tagging, \textit{NER}: Named Entity Recognition, \textit{SA}: Sentiment Analysis, \textit{SMT}: Statistical and \textit{NMT}: Neural Machine Translation, \textit{Intent-Clf}: Intent classification, \textit{Lang-ID}: Language identification, \textit{Emo-Regress}: Emotional regression. \textit{Wass.} denotes Wasserstein.}
    \label{tab:divergence-table}
\end{table*}

%%%%%%%%%%%%%%%%%%%%%%%%%%%%%%%%%%%%%%
%       Selecting Data               %
%%%%%%%%%%%%%%%%%%%%%%%%%%%%%%%%%%%%%%
\subsection{Data Selection}
\label{sec:data-selection}
Divergence measures are used to select a subset of text from the source domain that shares similar characteristics to the target domain.
\abhi{Since the source domain has labelled data,} the selected data serves as supervised data to train models in the target domain. \abhi{We note that % in the ensuing section, 
the literature pays closer attention to data selection for machine translation compared to other tasks. This can be attributed to its popularity in real-world applications and the difficulty of obtaining parallel sentences for every pair of language.}

%%%%%%%%%%%%%%%%%%%%%%%%%%%%%%%%%%%%%%
%       Simple word level features   %
%%%%%%%%%%%%%%%%%%%%%%%%%%%%%%%%%%%%%%
% MinCR: BUG I'm not clear of your overall argumentation in this paragraph.  I can't find a good reasoning for the points in this section, they seem a haphazard collection of random citations...  Can you restructure?
Simple word-level and surface-level text features like word and n-gram frequency distributions and $tf.idf$ weighted distributions have sufficient power to distinguish between text varieties and help in data selection. Geometric measures like cosine, used with word frequency distributions, are effective for selecting data in parsing and POS tagging \cite{plank-van-noord-2011-effective}.  \abhi{Instead of considering distributions as (sparse) vectors, one can get a better sense of the distance between distributions using information-theoretic measures.} \newcite{DBLP:conf/icdm/Remus12} find JS-Div effective for sentiment analysis. While 
% MinCR: what features are you referring to here?
% AbhiCR word level features. I made it explicit
word-level features are useful to select supervised data for an end-task, they also can be used to select data to pre-train language-models subsequently used for NER. \newcite{dai-etal-2019-using} use Term Vocabulary Overlap for selecting data for pretraining language models. 
Geometric and Information-theoretic measures with word level distributions are inexpensive to calculate. However, the distributions are sparse and continuous word distributions help in learning denser representations.

%%%%%%%%%%%%%%%%%%%%%%%%%%%%%%%%%%%%%%
%   Continuous representations for   %
%   Data selection                   %
%%%%%%%%%%%%%%%%%%%%%%%%%%%%%%%%%%%%%%
Continuous or distributed representations  of words, such as CBOW, Skip-gram \cite{mikolov13} and GloVe \cite{pennington-etal-2014-glove}, address shortcomings of representing text as sparse, frequency-based probability distributions by transforming them into dense vectors learned from free-form text. 
% MinCR: check -- added "e.g." -- ok?
% AbhiCR: Okay
A geometric measure (e.g., Word Vector Variance used with static word embeddings) is useful to select pre-training data for NER~\cite{dai-etal-2019-using}. Such selected data is found to be similar in \textit{tenor} (the participants in a discourse, the relationships between them, etc.) to the source data. 
\abhi{But static embeddings do not change according to the context of use. 
In contrast, contextual word representations (CWR) --- mostly derived from neural networks \cite{devlin-etal-2019-bert,peters-etal-2018-deep} --- capture contextual similarities between words in two domains. That is, the same word used in two domains in different contexts will have different embeddings. 
CWRs can be obtained from hidden representations of pretrained neural machine translation (NMT) models.
% MinCR BUG: this looks buggy -- CWR are CWR?
% AbhiCR: fixed
\cite{mccanncove} have found such representations along with P-norm effective for data selection in MT \cite{wang-etal-2017-sentence}.  Compared to representations from shallow NMT models, hidden representations of deep neural network language models (LM) like BERT have further improved data selection for NMT \cite{Aharoni2020UnsupervisedDC}. 
% MinCR: same ending note as last paragraph?
%AbhiCR: removed it even from the previous paragraph and added it att he end of the next paragraph 
}

%%%%%%%%%%%%%%%%%%%%%%%%%%%%%%%%%%%%%%
%   LMs and Cross entropy            %
%%%%%%%%%%%%%%%%%%%%%%%%%%%%%%%%%%%%%%
% \kmy{{\it How does a LM make decisions?  You just mean its probabilities?}} \abhi{{\it Yes, I will phrase this better}}
\abhi{Divergences can be measured by comparing the probabilities of a language model, in contrast to directly using its hidden representations. } If a LM trained on the target domain assigns high probability to a sentence from the source domain, then the sentence should have similar characteristics to the target domain. Cross Entropy (CE) between probability distributions from LMs capture this notion of similarity between two domains. They have been extensively used for data selection in statistical machine translation (SMT) \cite{yasuda-etal-2008-method,moore-lewis-2010-intelligent,axelrod-etal-2011-domain, duh-etal-2013-adaptation,liu-etal-2014-effective}. However, CE based methods for data selection are less effective for neural machine translation~\cite{Wees2017DynamicDS, silva-etal-2018-extracting}. Instead, 
% MinCR: use citet here see https://journals.aas.org/natbib/
\citet{Wees2017DynamicDS} come up with a dynamic subset selection where new subset is chosen every epoch during training. 
% MinCR: 3rd time again? Same ending note as last paragraph?
% But one has to note that sufficient amount of data should be available to train good language models in the target domain.
We note again the common refrain that sufficient amount of data should be available; here, to train good language models in the target domain.

%%%%%%%%%%%%%%%%%%%%%%%%%%%%%%%%%%%%%%
%   Classifier based measures        %
%%%%%%%%%%%%%%%%%%%%%%%%%%%%%%%%%%%%%%
Similar to language models, probabilistic scores from classifiers --- which distinguish between samples from two domains --- can aid data selection. The probabilities assigned by such classifiers in construing source domain text as target domain has been used as a divergence measures in machine translation~\cite{chen-huang-2016-semi}. However, the classifiers require supervised target domain data which is not always available. % Alternatively, instead of training domain classifier and then using it for data selection, 
% MinCR: again, as subject of sentence, use citet 
As an alternative, \citet{chen-etal-2017-cost} train a classifier and selector in an alternating optimisation manner.

%%%%%%%%%%%%%%%%%%%%%%%%%%%%%%%%%%%%%%
%   Findings                         %
%%%%%%%%%%%%%%%%%%%%%%%%%%%%%%%%%%%%%%
% MinCR too many ``different''s.  Not useful, replaced.
%AbhiCR: Thank yo ufor the change
From this literature review, we find that 
% different
distinct measures are effective for different NLP tasks.  \citet{ruder-plank-2017-learning} argue that owing to their varying task characteristics, different measures should apply. They show that learning a linear combination of % different 
measures is useful for NER, parsing and sentiment analysis. However, this is not always possible, especially in unsupervised domain adaptation where there is no supervised data in target domain.   We observe that information theoretic measures and geometric measures based on frequency distributions and continuous representations are common for text and structured prediction tasks (cf. \Cref{tab:divergence-table} in the 
% MinCR: global BUG: find and replace appendix for Appendix~X
appendix). The effectiveness of higher order measures for these tasks are yet to be ascertained.

Further, we find that for SMT data selection, variants of Cross Entropy (CE) measures are used extensively. However, the conclusions of \citet{Wees2017DynamicDS} are more measured regarding the benefits of CE and related measures for NMT. Contextual word representations with cosine similarity has found some initial exploration for neural machine translation (NMT), with higher order measures yet to be explored for data selection in NMT. 

%%%%%%%%%%%%%%%%%%%%%%%%%%%%%%%%%%%%%%
%       Learning Representations     %
%%%%%%%%%%%%%%%%%%%%%%%%%%%%%%%%%%%%%%

\subsection{Learning Representations}
\label{sec:learning-representations}

%%%%%%%%%%%%%%%%%%%%%%%%%%%%%%%%%%%%%%%%%%%%%%%%%%%%%%%%%%%%%%%%%%%%%%%%%%%%%%
% Introduction to DANN and DSN
%%%%%%%%%%%%%%%%%%%%%%%%%%%%%%%%%%%%%%%%%%%%%%%%%%%%%%%%%%%%%%%%%%%%%%%%%%%%%%
One way to achieve domain adaptation is to learn representations that are domain-invariant which are sufficiently powerful to perform well on an end task \cite{Ganin2015DomainAdversarialTO,ganinlempitsky}. The theory of domain divergence \cite{ben2010theory} shows that the target domain error is bounded by the source domain error and domain divergence ($\mathcal{H}$-divergence) and reducing the domain divergence results in domain-invariant representation. The theory also proposes a practical alternative to measure $\mathcal{H}$-divergence called PAD. The idea is to learn a representations that confuses a domain discriminator sufficiently to make samples from two domains indistinguishable.

% PAD, an approximation of $\mathcal{H}$-divergence, is large when a domain discriminator's error is small. Here, the discriminator \abhi{classifies source and target domain samples}. For domain-invariance, learned representations should not be able to separate source and target domain samples.

~\newcite{Ganin2015DomainAdversarialTO} operationalise %reducing 
PAD in a neural network named Domain Adversarial Neural Networks (DANN). The network employs a min--max game --- between the representation learner and the domain discriminator --- inspired by Generative Adversarial Networks \cite{gan}. The representation learner is not only trained to minimise a task loss on source domain, but also maximise a discriminator's loss, by reversing the gradients calculated for the discriminator. Note that this does not require any supervised data for target domain.  In later work,~\newcite{bousmalis} argue that domain-specific peculiarities are lost in a DANN, and propose \textit{Domain Separation Networks} (DSN) to address this shortcoming.  In DSN, both domain-specific and -invariant representations are captured in a \textit{shared--private} network. DSN is flexible in its choice of divergence measures and they find PAD performs better than MMD.
% Abhi: Moving it to just before starting this section's main part 
\abhi{Here, we limit our review to works utilising divergence measures. We exclude feature-based UDA methods such as Structural Corresponding Learning (SCL) \cite{blitzer-etal-2006-domain}, Autoencoder-SCL and pivot based language models \cite{ziser-reichart-2017-neural,ziser-reichart-2018-pivot,ziser-reichart-2019-task,bendavid2020perl}.}

%%%%%%%%%%%%%%%%%%%%%%%%%%%%%%%%%%%%%%%%%%%%%%%%%%%%%%%%%%%%%%%%%%
% Vanilla DANN and DSN works - Single source and target domain
%%%%%%%%%%%%%%%%%%%%%%%%%%%%%%%%%%%%%%%%%%%%%%%%%%%%%%%%%%%%%%%%%%%
 Obtaining domain invariant representations is desirable for many different NLP tasks, especially for sequence labelling where annotating large amounts of data is hard. They are typically used when there is a single source domain and a single target domain --- for sentiment analysis \cite{ganin2016}, NER \cite{zhou-etal-2019-dual}, stance detection \cite{Xu2019AdversarialDA}, machine translation \cite{britz-etal-2017-effective, zeng-etal-2018-multi}. The application of DANN and DSN to a variety of tasks are testament of their generality.
%%%%%%%%%%%%%%%%%%%%%%%%%%%%%%%%%%%%%%%%%%%%%%%%%%%%%%%%%%%%%%%%%%
% Vanilla DANN and DSN works - Considering some peculiarities 
% Incongruous text as different domains 
% Noisy text and clean text as two different domains 
%%%%%%%%%%%%%%%%%%%%%%%%%%%%%%%%%%%%%%%%%%%%%%%%%%%%%%%%%%%%%%%%%%%
 
 DANN and DSN are applied in other innovative situations. Text from two different periods of time can be considered as two different domains for intent classification \cite{kim-etal-2017-adversarial}.  \newcite{gui-etal-2017-part} consider clean formal newswire data as source domain and noisy, colloquial, unlabeled Twitter data as the target domain and use adversarial learning to learn robust representations for POS. Commonsense knowledge graphs can help in learning domain-invariant representations as well. \newcite{DBLP:conf/acl/GhosalHRMMP20} condition DANN with an external commonsense knowledge graph using graph convolutional neural networks for sentiment analysis. In contrast, \newcite{wang-etal-2018-label} use MMD outside the adversarial learning framework. They use MMD to learn to reduce the discrepancy between neural network representations belonging to two domains. Such concepts have been explored in computer vision \cite{tzengetalddc}.

%%%%%%%%%%%%%%%%%%%%%%%%%%%%%%%%%%%%%%%%%%%%%%%%%%%%%%%%%%%%%%%%%%
% Multiple source and single target domain extension 
%%%%%%%%%%%%%%%%%%%%%%%%%%%%%%%%%%%%%%%%%%%%%%%%%%%%%%%%%%%%%%%%%%%
While single source and target domains are common, complementary information available in multiple domains can help to improve performance in a target domain. This is especially helpful when there is no large-scale labelled data in any one domain, but where smaller amounts are available in several domains. DANN and DSN have been extended to such multi-source domain adaptation: for intent classification \cite{dingmultidomain}, sentiment analysis \cite{chen-cardie-2018-multinomial, li-etal-2018-whats, guo-etal-2018-multi, wright2020transformer} and machine translation \cite{gu-etal-2019-improving, Wang2019GoFT}.

%%%%%%%%%%%%%%%%%%%%%%%%%%%%%%%%%%%%%%%%%%%%%%%%%%%%%%%%%%%%%%%%%%
% Adversarial learning for multi-task learning
%%%%%%%%%%%%%%%%%%%%%%%%%%%%%%%%%%%%%%%%%%%%%%%%%%%%%%%%%%%%%%%%%%%
DANN and DSN can also help in multitask learning which considers two complementary tasks \cite{Caruana97}. A key to multitask learning is to learn a shared representation that captures the common features of two tasks. However, such representations might still contain task-specific information. The shared-private model of DSN helps in disentangling such representations and has been used for sentiment analysis \cite{liu-etal-2017-adversarial}, Chinese NER and word segmentation \cite{cao-etal-2018-adversarial}.
%%%%%%%%%%%%%%%%%%%%%%%%%%%%%%%%%%%%%%%%%%%%%%%%%%%%%%%%%%%%%%%%%%
% Adversarial learning for multilingual works
%%%%%%%%%%%%%%%%%%%%%%%%%%%%%%%%%%%%%%%%%%%%%%%%%%%%%%%%%%%%%%%%%%%
Also, although beyond the scope of our discussion here, DANN and DSN have been used to learn language-agnostic representations for text classification and structured prediction in multilingual learning \cite{chen-etal-2018-adversarial, zou-etal-2018-adversarial,yasunaga-etal-2018-robust}.

%%%%%%%%%%%%%%%%%%%%%%%%%%%%%%%%%%%%%%%%%%%%%%%%%%%%%%%%%%%%%%%%%%
% Works trying different other measures for stable training 
%%%%%%%%%%%%%%%%%%%%%%%%%%%%%%%%%%%%%%%%%%%%%%%%%%%%%%%%%%%%%%%%%%%
Most works that adopt DANN and DSN framework reduce either the PAD or MMD divergence. However, reducing the divergences, combined with other auxiliary task specific loss functions, can result in training instabilities and vanishing gradients when the domain discriminator becomes increasingly accurate \cite{shenetal2018}.  Using other higher order measures can result in more stable learning. In this vein, CMD has been used for sentiment analysis \cite{Zellinger2017CentralMD, peng-etal-2018-cross}, and Wasserstein distance has been used for duplicate question detection \cite{shah-etal-2018-adversarial} and to learn domain-invariant attention distributions for emotional regression \cite{zhu-etal-2019-adversarial}.

%%%%%%%%%%%%%%%%%%%%%%%%%%%%%%%%%%%%%%%%%%%%%%%%%%%%%%%%%%%%%%%%%%
% Conclusion
%%%%%%%%%%%%%%%%%%%%%%%%%%%%%%%%%%%%%%%%%%%%%%%%%%%%%%%%%%%%%%%%%%%
% \kmy{{\it Generally I have a problem with a paper whose important artefacts are self-contained within the page limit. Your tables fall much farther back and into the supplemental text.  Try to push Table 2 into the main text.  Also, why bold for the two statements here?}} \abhi{\it{I would like the table in the main part of the paper as well. I won't be left with any more space then. Unfortunately, the table takes a whole page. May be in the final publishable version of the paper, if they allow one more page then I can add it}}
The review shows that most works extend the DSN framework to learn domain invariant representations in different scenarios (cf. Table~\ref{tab:divergence-table}, in the appendix). The original work from \cite{bousmalis} includes MMD divergence besides PAD, which is not adopted in subsequent works, possibly due to the reported poor performance. Most works require careful balancing between multiple objective functions \cite{han-eisenstein-2019-unsupervised}, which can affect the stability of training. The stability of training can be improved by selecting appropriate divergence measures like CMD \cite{Zellinger2017CentralMD} and Wasserstein Distance \cite{wgan}. We believe additional future works will adopt such measures.

%%%%%%%%%%%%%%%%%%%%%%%%%%%%%%%%%%%%%%
%       Applications in the Wild     %
%%%%%%%%%%%%%%%%%%%%%%%%%%%%%%%%%%%%%%

\subsection{Decisions in the Wild}
\label{sec:applications-in-the-wild}
Models can perform poorly when they are deployed in the real world. The performance degrades due to the difference in distribution between training and test data. Such performance degradation can be alleviated by large-scale annotation in the new domain. However, annotation is  expensive, and --- given thousands of domains --- quickly becomes infeasible. Predicting the performance in a new domain, where there is no labelled data, is thus important. Much recent work provides theory \cite{DBLP:conf/iclr/RosenfeldRBS20, DBLP:journals/corr/chuang2020icml,Steinhardt2016UnsupervisedRE}. As models are put into production in the real world, this application becomes practically important as well. Empirically, NLP considers the divergence between the source and the target domain to predict performance drops.

% \subsubsection{Text Classification and Structured Prediction}

Simple measures based on word level features have been used to predict the performance of a machine learning model in new domains. Information theoretic measures like Renyi-Div and KL-Div has been used for predicting performance drops in POS \cite{van-asch-daelemans-2010-using} and Cross-Entropy based measure has been used for dependency parsing \cite{ravi-etal-2008-automatic}.  Prediction of performance can also be useful for machine translation where obtaining parallel data is hard. Based on distance between languages, \cite{xia2020predicting} predict performance of the model on new languages for MT, among other tasks.  Such performance prediction models have also been done in the past for SMT \cite{birch-etal-2008-predicting, specia-etal-2013-quest}.
However, \citet{ponomareva2012biographies} argue that predicting {\it drops in performance} is more appropriate compared to raw performance. They find that JS-Div effective for predicting performance drop of Sentiment Analysis systems. 

Only recently, predicting model failures in practical deployments from an empirical viewpoint has regained attention. \citet{elsahar-galle-2019-annotate} find the efficacy of higher-order measures to predict the drop in performance for POS and SA and do not rely on hand crafted measures as in previous works. 
% Min: on what basis can you say this?
% \kmy{{\it Not sure you need the bold for this.}} \abhi{\textit{Just a highlight on what is lacking in the literature. Can be removed}}
However, analysing performance drops using CWR is still lacking. We tackle this in the next section.

% Apart from the aforementioned NLP tasks, information-theoretic measures like JS-Div are also used to detect the change in the meaning of words across time \cite{schlechtweg-etal-2019-wind} and to capture the divergence between two languages \cite{asgari-mofrad-2016-comparing}. 

% To summarise, a multitude of domain divergence measures are used across applications and tasks. While information-theoretic and geometric measures have found widespread use, the effectiveness of using higher-order measures and continuous representations is still unclear. We further explore this by providing a comprehensive analysis next.

% % **********************************
% %       EMPIRICAL ANALYSIS
% % **********************************
\section{Experiments}
\label{sec:empirical-analysis}

\begin{figure*}
    \subfloat[(POS)-MMD-Gaussian\label{fig:pos-mmd-gaussian}]{{\includegraphics[width=0.31\linewidth]{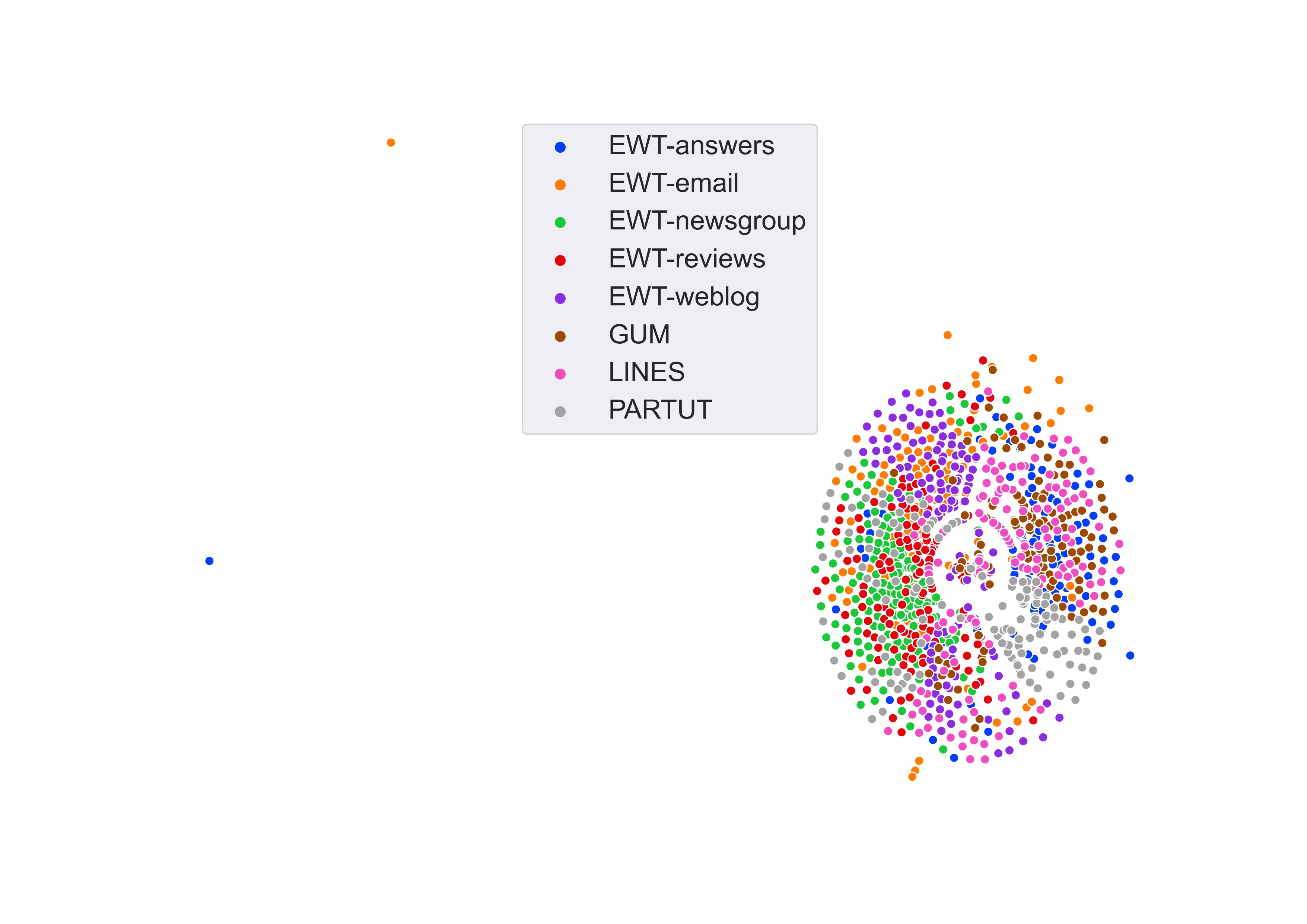} }}%
    \subfloat[(NER)-MMD-RQ\label{fig:ner-cos}]{{\includegraphics[width=0.31\linewidth]{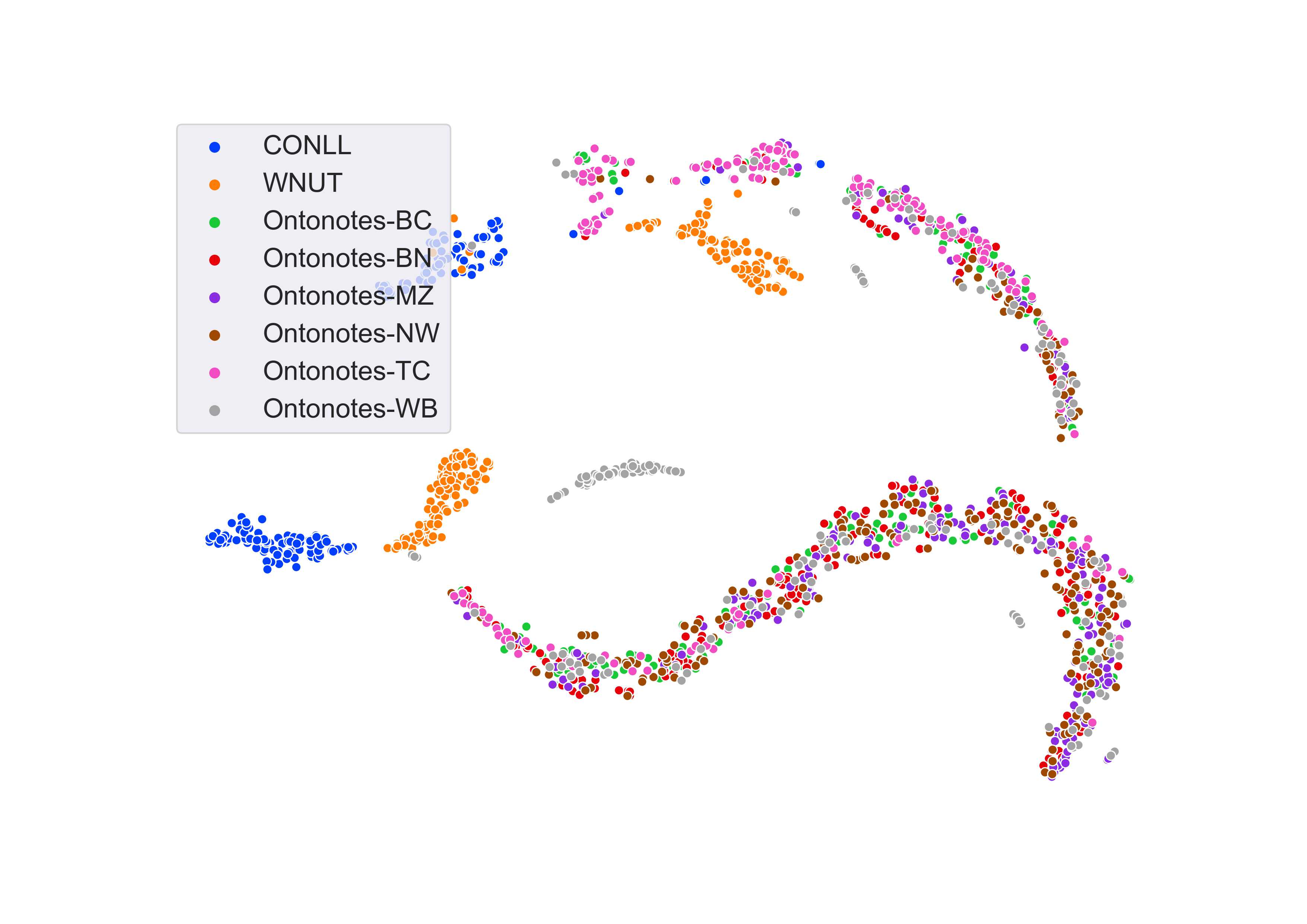} }}%
    \subfloat[(SA)-JS-Div\label{fig:sa-js-div}]{{\includegraphics[width=0.31\linewidth]{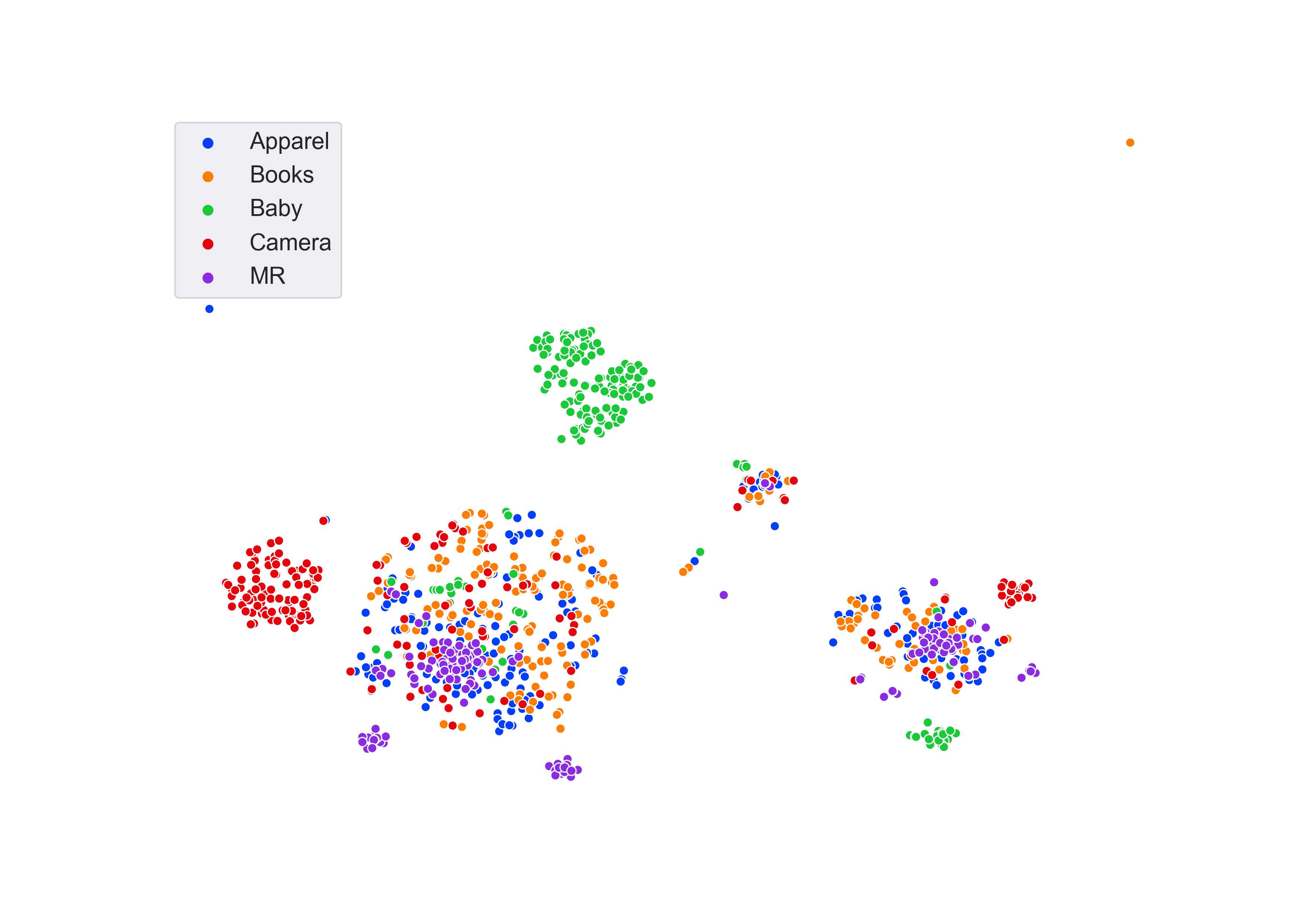} }}%
    \label{fig:silhouette-analysis}%
    \caption{t-SNE plots for select measures. The complete set of diagrams are available in Appendix \ref{sec:tsne-plots}.}
\end{figure*}

A practical use case of domain divergences is to predict the performance drop of a model applied to a new domain. We ask how relevant are traditional measures over word distributions compared to higher-order measures like CMD and MMD over contextual word representations like BERT, Elmo, DistilBERT \cite{devlin-etal-2019-bert, peters-etal-2018-deep, Sanh2019DistilBERTAD}?
% that have become popular.
We perform an empirical study to assess their suitability to predict performance drops for three important NLP tasks: POS, NER, and SA leaving machine translation to future work.

Performance difference between the source and the target domain depends on the divergence between their feature distributions \cite{ben2010theory}. We assume a co-variate shift, as in \cite{ganin2016}, where the marginal distribution over features change, but the conditional label distributions does not --- i.e., $P_{\mathcal{D}_{s}}(y|x) = P_{\mathcal{D}_T}(y|x)$ $\mathcal{P}_{\mathcal{D}_{s}}(x) \neq P_{\mathcal{D}_T}(x)$. Although difference in conditional label distribution can increase the $\mathcal{H}$-Divergence measure \cite{wisniewski-yvon-2019-bad}, it requires labels in the target domain for assessment. In this work, we assume no labelled data in the target domain, to best mimic realistic settings.

\subsection{Experimental Setup}
{\bf Datasets:}
For POS, we select 5 different corpora from the English Word Tree Bank of Universal Dependency corpus \cite{nivre-etal-2016-universal}\footnote{Yahoo! Answers, Email, NewsGroups, Reviews and Weblogs.} and also include the GUM, Lines, and ParTUT datasets.  We follow \newcite{elsahar-galle-2019-annotate} and consider these as 8 domains. For NER, we consider CONLL 2003 \cite{tjong-kim-sang-de-meulder-2003-introduction}, Emerging and Rare Entity Recognition Twitter  \cite{derczynski-etal-2017-results} and all 6 categories in OntoNotes~v5 \cite{hovy-etal-2006-ontonotes}\footnote{Broadcast News (BN), Broadcast Conversation (BC), Magazine (MZ), Telephone Conversation (TC) and Web (WB).}, resulting in 8 domains. For SA, we follow \newcite{guo2020multisource}, selecting % Min3: what is MR?
the same 5 categories\footnote{Apparel, Baby, Books, Camera and MR.} for experiments \cite{liu-etal-2017-adversarial}. 

% \subsubsection{
\noindent
{\bf Divergence Measures:}
We consider 12 divergences. For Cos, we follow the instance based calculation \cite{Ruder2017DataSS}. For MMD, Wasserstein and CORAL, we randomly sample 1000 sentences and average the results over 3 runs. For MMD, we experiment with different kernels (\textit{cf.} ~\Cref{sec:background}) and use default values of $\sigma$ from the GeomLoss package \cite{feydy2019interpolating}.
For TVO, KL-div, JS-div, Renyi-div, based on word frequency distribution we remove stop-words and consider the top 10k frequent words across domains to build our vocabulary \cite{Ruder2017DataSS,gururangan2020dont}. We use
% Min3: then isn't Renyi-div then just KL-div?
$\alpha$\textit{=0.99} for Renyi as found effective by~\newcite{plank-van-noord-2011-effective}. 
We do not choose CE as it is mainly used in MT and 
% Min3: can rewrite as "ineffective for text classification ..."
ineffective for classification and structured prediction~\cite{Ruder2017DataSS}.

% \subsubsection{
\noindent 
{\bf Model Architecture:}
For all our experiments, unless otherwise mentioned, we use the pre-trained DistilBERT~\cite{Sanh2019DistilBERTAD} model. It has competitive performance to BERT, but has faster inference times and lower resource requirements. 
% MinCR: maybe not needed?
% We leave experimentation with other BERT variants, such as Roberta~\cite{Liu2019RoBERTaAR}, for future work. 
%AbhiCR: This was a defending statement for the reviewer, since Roberta was getting famous at that point in time. We can leave this out.
For every text segment, we obtain the activations from the final layer and average-pool the representations. We train the models on the source domain training split and test the best model --- picked from validation set grid search --- on the test dataset of the same and other domains (cf.~\Cref{sec:cross-domain-perfs}).  

For POS and NER, we follow the original BERT model where a linear layer is added and a prediction is made for every token. If the token is split into multiple tokens due to Byte Pair Encoding, the label for the first token is predicted. For SA and domain discriminators, we pool the representation from the last layer of DistilBERT and add a linear layer for prediction (\Cref{sec:hyperparams}).

\begin{table*}
    \small 
    \centering
    \begin{tabular}{p{2.15cm}|ccc|ccc}
    \hline 
    \textbf{Measure} & \multicolumn{3}{|c|}{\textbf{Correlations}} & \multicolumn{3}{c}{\textbf{Silhouette Coefficients}} \\ 
    \hline 
    - & \textbf{POS} & \textbf{NER} & \textbf{SA} & \textbf{POS} & \textbf{NER} & \textbf{SA} \\
    \hline 
    \rowcolor{Geometric}
    Cos & 0.018  & 0.223 & -0.012 & $-1.78 \times 10^{-1}$  & $-2.49\times 10^{-1}$ & $-2.01 \times 10^{-1}$ \\
    \rowcolor{InformationTheoretic}
    KL-Div & 0.394  & 0.384 & 0.715 & - & -  & - \\
    \rowcolor{InformationTheoretic}
    JS-Div &  0.407  & 0.484  & 0.709 & $-8.50 \times 10^{-2}$ & $-6.40 \times 10^{-2}$ & $+2.04 \times 10^{-2}$ \\
    \rowcolor{InformationTheoretic}
    Renyi-Div &  0.392   & 0.382  & \textbf{0.716} & - & -  & - \\
    \rowcolor{HigherOrder}
    PAD & \textbf{0.477} & 0.426 & 0.538 & - & - & -  \\
    \rowcolor{HigherOrder}
    Wasserstein & 0.378 & 0.463  & 0.448 & $-2.11 \times 10^{-1}$ & $-2.36 \times 10^{-1}$ &  $-1.70\times 10^{-1}$ \\
    \rowcolor{HigherOrder}
    MMD-RQ &  0.248  & \textbf{0.495} & 0.614 & $-4.11 \times 10^{-2}$ & $-3.04 \times 10^{-2}$ & $-1.70\times10^{-2}$ \\
    \rowcolor{HigherOrder}
    MMD-Gaussian & 0.402 & 0.221  & 0.543 & $+4.25 \times 10^{-5}$ & $+2.37\times 10^{-3}$ & $-8.42\times 10^{-5}$ \\
    \rowcolor{HigherOrder}
    MMD-Energy & 0.244 & 0.447 & 0.521 & $-9.84 \times 10^{-2}$  & $-1.14 \times 10^{-1}$ & $-8.48 \times 10^{-2}$ \\
    \rowcolor{HigherOrder}
    MMD-Laplacian & 0.389 & 0.273  & 0.623 & $-1.67 \times 10^{-3}$ & $+4.26 \times 10^{-4}$ & $-1.08 \times 10^{-3} $ \\
    \rowcolor{HigherOrder}
    CORAL & 0.349 & 0.484   &  0.267 & $-2.34 \times 10^{-1}$ & $-2.78 \times 10^{-1}$ & $-1.41 \times 10^{-1}$ \\
    \rowcolor{Gray1}
    TVO &-0.437 & -0.457 & -0.568  & - & - & - \\ 
    \hline 
    \end{tabular}
    \caption{(l): Correlation of performance drops with divergence measures. Measures with higher correlations are better indicators of performance drops. (r): Silhouette coefficients considering different divergence measures. We randomly sample 200 points for calculation and average the results over 5 runs. Only certain divergences which are metrics are allowed. The colours are from the taxonomy of divergence measures in \Cref{fig:domain-adaptation-measures}.}
    \label{tab:corr-silhouette}
\end{table*}

\subsection{Are traditional measures still relevant?}

For POS, the PAD measure has the best correlation with performance drop (cf. Table \ref{tab:corr-silhouette}).  Information-theoretic measures over word frequency distributions, such as JS-div, KL-div, and TVO, which have been prevalent for data selection and performance drop use cases (cf. Table \ref{tab:divergence-table}) are comparable to PAD.  \newcite{plank-etal-2014-importance} claim that the errors in POS are dictated by out of vocabulary words.  % We find that the high correlations among KL-div, JS-div, and TVO --- all based on word probability distributions --- with drop in performance corroborate their claims. 
\kmy{Our findings validate their claim, as we find strong correlation between POS performance drop and word probability distribution measures}
% MinCR: don't need again.
% (KL-div, JS-div and TVO). 
For NER, MMD-RQ provides the best correlation of 0.495. CORAL --- a higher-order measure --- and JS-div are comparable. For SA, Renyi-div and other information-theoretic measures provide considerably better correlation compared to higher-order measures. 
% Highlighting  on individual measures and their relation to other works in the literature 
Cos is a widely-used measure across applications, however it \textit{did not} provide significant correlation for either task. TVO is used for selecting pretraining data for NER \cite{dai-etal-2019-using} and as a measure to gauge the benefits of fine-tuning pre-trained LMs on domain-specific data \cite{gururangan2020dont}. Although TVO does not capture the nuances of domain divergences, it has strong, reliable correlations for performance drops. PAD has been suggested for data selection in SA by \newcite{ruder-plank-2017-learning} and for predicting drop in performance by \newcite{elsahar-galle-2019-annotate}. Our analysis confirms that PAD provides good correlations across POS, NER, and SA. 

We find no single measure to be superior across all tasks. However, information theoretic measures consistently provide good correlations. 
Currently, when contextual word representations dictate results in NLP, simple measures based on frequency distributions are strong baselines for predicting performance drop.
Although higher-order measures do not always provide the best correlation, they are differentiable, thus suited for end-to-end training of domain-invariant representations. 

\subsection{Discussion}

Why are some divergence measures better at predicting drops in performance?  The {\it one-dataset-one-domain} is a key assumption in such works.
% {\it one-dataset-is-one-domain}
% {\it dataset-is-domain}
% different datasets are different domains. 
However, many works have questioned this assumption \cite{plank-van-noord-2011-effective}. Multiple domains may exist within the same domain \cite{webber-2009-genre} and two different datasets may not necessarily be considered different domains \cite{irvine-etal-2013-measuring}.  Recently \newcite{Aharoni2020UnsupervisedDC} show that BERT representations reveal their underlying domains. They qualitatively show that a few text segments from a dataset 
% MinCR: why ``aptly''?
% AbhiCR: I meant "actually"... they were supposed to belong to the other domain. They actually belong to the other domain but included in this dataset. Changing it
actually belong to another domain. However the degree to which the samples belong to different domains is unclear.

We first test the assumption that different datasets are different domains using Silhouette scores~\cite{rousseeuw1987} which quantify the separability of clusters. We initially assume that a dataset is in its own domain. A positive score shows that datasets can be considered as well-separated domains; a negative score shows that most of the points within a dataset can be assigned to a nearby domain; and 0 signifies overlapping domains. \abhi{We calculate Silhouette scores and t-SNE plots \cite{maaten2008visualizing} for different divergence measures. Refer to the plots (\Cref{fig:pos-tsne,fig:ner-tsne,fig:sa-tsne}) and calculation details in \Cref{sec:tsne-plots}. }

% For calculations we use a subset of domain divergence measures that are metrics (a requirement of Silhouette scores) and can be calculated between single instances of text. We sample 200 points for each dataset as the time complexity increases exponentially with number of points.  We average the results over 5 runs. We further obtained t-SNE plots \cite{maaten2008visualizing} for different divergence measures with DistilBERT representations (\Cref{fig:pos-tsne,fig:ner-tsne,fig:sa-tsne}, of \Cref{sec:tsne-plots}).

Almost all the measures across different tasks have negative values close to 0 (Table~\ref{tab:corr-silhouette}, (r)).

\vspace{-2.1mm}
\begin{itemize}[leftmargin=*]
    \itemsep-0.4em
    \item For POS, CORAL, Wasserstein and Cos strongly indicate that text within a dataset belongs to other domains. However, for MMD-Gaussian \abhi{the domains overlap (\Cref{fig:pos-mmd-gaussian})}. 
    \item For NER, MMD-Gaussian and MMD-Laplacian indicate that the clusters overlap while all other metrics have negative values.
    \item For SA, JS-Div has positive values compared to other measures, and as seen in Figure~\ref{fig:sa-js-div}, we can see a better notion of distinct clusters.
\end{itemize}
\vspace{-2mm}

% POS -- CORAL, Wasserstein, and Cos strongly indicate that text within a dataset belongs to other domains. However, for MMD-Gaussian, as seen in~\Cref{fig:pos-mmd-gaussian}, we do observe some clustering, but the domains are overlapping. For NER, MMD-Gaussian and MMD-Laplacian indicate that the clusters overlap while all other metrics have negative values. For sentiment analysis, JS-div has positive values compared to other measures, and as seen in Figure~\ref{fig:sa-js-div}, we can see a better notion of distinct clusters. 

The Silhouette scores along with the t-SNE plots show that datasets are, in fact, not distinct domains.  Considering data-driven methods for defining domains is needed~\cite{Aharoni2020UnsupervisedDC}.

If there are indeed separate domains, does it explain why some measures are better than the others? We see better notions of clusters for NER and sentiment analysis ({\it cf.} Figures~\ref{fig:ner-cos} and \ref{fig:sa-js-div}).  We can expect the drop in performance to be indicative of these domain separations. Comparing the best correlations from Table~\ref{tab:corr-silhouette}, correlations for NER and sentiment analysis are higher compared with POS.  For POS, there are no indicative domain clusters and the correlation between domain divergence and performance may be less; whereas for SA, both the t-SNE plot and the Silhouette scores for JS-Div ({\it cf.} Figure~\ref{fig:sa-js-div}) corroborate comparatively better separation. \abhi{If datasets are indeed different domains, these divergence measures are reliable indicators of performance drops. If they are not, there might be other confounding factors (such as differences in label distribution) and one has to be cautious in using them.}

\abhi{Domain overlap} also has consequences for data selection strategies. For example, \newcite{moore-lewis-2010-intelligent} select \textit{pseudo in-domain data} from source corpora (cf \Cref{sec:data-selection}). As the Silhouette coefficients are negative and close to 0, many data points in a dataset belong to nearby domains. Data selection strategies thus may be effective. If the Silhouette coefficients are more negative and if more points in the source aptly belong to the target domain, we should expect increased sampling from such source domains to yield additional performance benefits in the target domain.

% % **********************************
% %           CONCLUSION
% % **********************************
\section{Conclusion}
\label{sec:conclusion}
We survey domain adaptation works, focusing on divergence measures and their usage for \textit{data selection}, \textit{learning domain-invariant representations}, and \textit{making decisions in the wild}. We synthesised the divergence measures into a taxonomy of \textit{information theoretic}, \textit{geometric} and \textit{higher-order} measures. While traditional measures are common for data selection and making decisions in the wild, higher-order measures are prevalent in learning representations. 
% The correlation analysis showed that higher-order measures with CWRs need not be the best indicators of performance drops but they have properties that make them amenable for learning domain-invariant representations using neural networks, while traditional measures still serve as strong baselines. 
% Min2: Your conclusion needs to be stronger, but even now I don't know where the basis for support is for these recommendations.
% Abhi2: Removing this for now. Even I am uncomfortable making this recommendation.
% We recommend using information theoretic measures when there is significant amount of data to estimate probability distributions reliably, and using higher-order measures over CWRs in extremely low-resource setting.  
% Our experimental results using Silhouette scores and t-SNE plots indicate that datasets need not be considered as separate domains. We observe that in cases when the domains are separated, the correlations for performance drops are higher. Different datasets should not be treated as domain without subjecting them to analysis. We expect that future works define domains in a data-driven manner for their experiments for POS, NER and SA.
Based on our correlation experiments, silhouette scores, and t-SNE plots, we make the following recommendations:

\vspace{-2.4mm}
\begin{itemize}[leftmargin=*]
\itemsep-0.37em 
\item  PAD is a reliable indicator of performance drop. It is best used when there are sufficient examples to train a domain discriminator.
\item JS-Div is symmetric and a formal metric. It is related to PAD, easy to compute, and serves as a strong baseline.
\item While Cosine is popular, it is an unreliable indicator of performance drop.
\item One-dataset-is-not-one-domain. Instead, cluster representations and define appropriate domains.
\end{itemize}

\clearpage

\section*{Acknowldgements}
We  would  also  like  to  acknowledge the support of the NExT research grant funds,supported  by  the  National  Research  Foundation,Prime Ministers Office,  Singapore under its IRC@  SG  Funding  Initiative,  and  to  gratefully  acknowledge  the  support  of  NVIDIA  Corporation  with  the  donation  of  the  GeForce  GTX  Titan  XGPU used in this research.

\bibliography{naaclhlt2019}
\bibliographystyle{acl_natbib}

\clearpage
\appendix
% **********************************
%           BACKGROUND
% **********************************
% \input{sections/divergence_table}
\section{Domain Divergence Measures}
\label{sec:background}

This section provides the necessary background on different kinds of divergence measures used in the literature. They can be either information-theoretic -- which measure the distance between two probability distributions, geometric - which measure the distance between two vectors in a space, or higher-order which capture similarity in a projected space and consider higher order moments of random variables.

%%%%%%%%%%%%%%%%%%%%%%%%%%%%%%%%%
% KL Divergence                 %
%%%%%%%%%%%%%%%%%%%%%%%%%%%%%%%%%
\subsection{Information-Theoretic Measures}
Let $P$ and $Q$ be two probability distributions. These information-theoretic measures are used to capture differences between $P$ and $Q$. \\

\noindent 
\textbf{Kullback-Leibler Divergence (KL-Div)} $Q$ is called the reference probability distribution\footnote{KL divergence is asymmetric and cannot be considered a metric}. More precisely, KL is defined if only for all $Q(x)$ st $Q(x) = 0$, $P(x)$ is also 0; and undefined if $\exists$ x, $Q(x)=0$ and $P(x) > 0$.

\begin{equation}
    D_{KL}(P || Q) = \sum_{x} P(x) log\biggl(\frac{P(x)}{Q(x)} \biggr)
\end{equation}

%%%%%%%%%%%%%%%%%%%%%%%%%%%%%%%%%
% Renyi Divergence              %
%%%%%%%%%%%%%%%%%%%%%%%%%%%%%%%%%
\noindent
\textbf{Renyi Divergence (Renyi-Div)} Renyi Divergence is a generalisation of the KL Divergence and is also called $\alpha$-power divergence:

\begin{equation}
    D_{\alpha}(P||Q) = \frac{1}{\alpha - 1} log \biggl( \sum_{x}  \frac{P(x)^{\alpha}}{Q(x)^{\alpha - 1}}\biggr)
\end{equation}

Here $\alpha \geq 0$ and $\alpha \ne 1$. Renyi divergence is equivalent to KL divergence in the limit where $\alpha \rightarrow 1$.\\

%%%%%%%%%%%%%%%%%%%%%%%%%%%%%%%%%
%               JSD             %
%%%%%%%%%%%%%%%%%%%%%%%%%%%%%%%%%
\noindent 
\textbf{Jensen Shannon Divergence (JS-Div)} Jensen Shannon divergence (JS-divergence) is a symmetric version of KL-Divergence. It has many advantages. The square root of the Jensen Shannon Divergence is a metric and it can be used for non-continuous probabilities:

\begin{equation}
    \begin{split}
    D_{JS}(P || Q) = \frac{1}{2} D_{KL}(P || M) + \frac{1}{2}D_{KL}(Q || M) \\
    M = \frac{1}{2} (P + Q)
    \end{split}
\end{equation}

\noindent 
\textbf{Entropy-Related - (CE)}  Let, $H_T$, $H_S$ assign entropy to a sentence using a language model trained on the target and source domain, respectively. If $s$ is a text segment from the source domain, then the difference in entropy, as shown below, gives the similarity of a source domain segment to the target domain. Some works just use $H_T$, ignoring $H_S$. MT related work \cite{moore-lewis-2010-intelligent}, consider only the source language. 
% MinCR: looks dangling.  Fix sentence structure.
% AbhiCR: Fixed it
\newcite{axelrod-etal-2011-domain} extend to consider both the source and the target language of machine translation, which performs better for data selection. We present these variations in the formulae below and attribute the same name \textbf{CE} to both these variations in the literature review.

\begin{equation}
    D_{CE} = H_T(s) - H_S(s)
\end{equation}

\begin{equation}
    \begin{aligned}
    D_{CE} = [H_T^{src-lang}(s) - H_S^{src-lang}(s)] \\ +  [H_T^{trg-lang}(s) - H_S^{trg-lang}(s)]
    \end{aligned}
\end{equation}

%%%%%%%%%%%%%%%%%%%%%%%%%%%%%%%%%
% Cosine Similarity             %
%%%%%%%%%%%%%%%%%%%%%%%%%%%%%%%%%
\subsection{Geometric Measures}
Let $\vec{p}$ and $\vec{q}$ be two vectors in $\mathbb{R}^{n}$. Domain adaptation works use geometric metrics for continuous representations like word vectors. \\

\noindent
\textbf{Cosine Similarity (Cos)}: It calculates the cosine of the angle between vectors. To measure the cosine distance between two points, we use $1-Cos$:

\begin{equation}
    cos(\vec{p}, \vec{q}) = \frac{\vec{p}. \vec{q}}{\norm{p}. \norm{q}}
\end{equation}

\noindent 
\textbf{$l_p$-norm (Norm)}: 
\textbf{Euclidean distance or $l_2$ distance} measures the straight line distance between vectors and \textbf{Manhattan or $l_1$} measures the sum of the difference between their projections.

\begin{gather}
d_2(p, q) = \sqrt{\sum_{i=1}^{n} (p_i - q_i)^2} \\
d_1(p, q) = \sum_{i=1}^{n} |p_i - q_i|
\end{gather}

%%%%%%%%%%%%%%%%%%%%%%%%%%%%%%%%%
% PAD              %
%%%%%%%%%%%%%%%%%%%%%%%%%%%%%%%%%
\subsection{Higher-Order Measures}
\noindent
\textbf{$\mathcal{H}$-divergence and Proxy-A-Distance (PAD): }
\newcite{ben2010theory} state that the error of a machine learning classifier in a target domain is bound by its performance on the source domain and the $\mathcal{H}$-divergence between the source and the target distributions. $\mathcal{H}$-divergence is expensive to calculate.  An approximation of $\mathcal{H}$ is called \textit{Proxy-A-Distance}. This definition has been adopted from \cite{elsahar-galle-2019-annotate}. Here $G: \mathcal{X} \rightarrow [0, 1]$ is a supervised machine learning model that classifies examples to the source and target domains, $D_s, D_t$. $|D|$ is the size of the training data and $\mathbbm{1}$ is an indicator function:

    \begin{gather}
     PAD = 1 - 2\epsilon(G_d) \\ 
     \epsilon(G_d) = 1 - \frac{1}{|D|} \sum_{x_i \in D_s, D_t}|G(x_i) - \mathbbm{1}(x_i \in D_s)|
    \end{gather}

%%%%%%%%%%%%%%%%%%%%%%%%%%%%%%%%%
%      Wasserstein Distance     %
%%%%%%%%%%%%%%%%%%%%%%%%%%%%%%%%%

\noindent 
\textbf{Wasserstein Distance: }
Wasserstein Distance (also called Earth Mover's distance) is another metric for two probability distributions. Intuitively, it measures the least amount of work done to transport probability mass from one probability distribution to another to make them equal. The work done in this case is measured as the mass transported multiplied by the distance of travel. It is known to be better than Kullback-Leibler Divergence and Jensen-Shannon Divergence when the random variables are high dimensional or otherwise. The Wasserstein metric is defined as:

\begin{equation*}
    D_{Wasserstein} = \underset{\gamma \in \pi}{inf} \sum_{x, y} \norm{x-y} \gamma(x, y)
\end{equation*}

Here $\gamma \in \pi(P, Q)$ where $\pi(P, Q)$ is the set of all distributions where the marginals are $P$ and $Q$. \\

%%%%%%%%%%%%%%%%%%%%%%%%%%%%%%%%%
%           MMD                 %
%%%%%%%%%%%%%%%%%%%%%%%%%%%%%%%%%
\noindent
\textbf{Maximum Mean Discrepancy (MMD)}: MMD is a non-parametric method to estimate the distance between distributions based on Reproducing Kernel Hilbert Spaces (RKHS). Given two random variables $X = \{x_1, x_2, ... ,x_m\}$ and $Y=\{y_1, y_2, ...., y_n\}$ that are drawn from distributions $P$ and $Q$, the empirical estimate of the distance between distribution $P$ and $Q$ is given by:

\begin{equation}
    \small
    MMD(X,Y) = \norm{ \frac{1}{m} \sum_{i=1}^{m} \phi(x_i) - \frac{1}{n} \sum_{i=1}^{n} \phi(y_i) }_{\mathcal{H}}
\end{equation}

Here $\phi: \mathcal{X} \rightarrow \mathcal{H}$ are nonlinear mappings of the samples to a feature representation in a RKHS. In this work, we map the contextual word representations of the text to RKHS. The different kinds of kernels we use in this work are given below. We use the default values of $\sigma=0.05$ of the GeomLoss package \cite{feydy2019interpolating}. 

\textbf{Rational Quadratic Kernel}
\begin{equation*}
    \phi(x, y) = \bigg( 1 + \frac{1}{2\alpha} (x-y) ^ T \Theta^{-2} (x -y) \bigg)^{-\alpha}
\end{equation*}

\textbf{Energy}
\begin{equation*}
    \phi(x, y) = -\norm{x-y}_2
\end{equation*}

\textbf{Gaussian}
\begin{equation*}
    \phi(x, y) = exp(-\frac{\norm{x-y}_2^2}{2 \sigma^2})
\end{equation*}

\textbf{Laplacian}
\begin{equation*}
    \phi(x, y) = exp(-\frac{\norm{x-y}_2}{\sigma})
\end{equation*}

%%%%%%%%%%%%%%%%%%%%%%%%%%%%%%%%%
%           CORAL                 %
%%%%%%%%%%%%%%%%%%%%%%%%%%%%%%%%%
\noindent 
\textbf{Correlation Alignment (CORAL)}:
Correlation alignment is the distance between the second-order moment of the source and target samples. If $d$ is the representation dimension, $\norm{}_F$ represents Frobenius norm and $Cov_S, Cov_T$ is the covariance matrix of the source and target samples, then CORAL is defined as:

\begin{equation}
    D_{CORAL} = \frac{1}{4d^2} \norm{Cov_S - Cov_T}_F^{2}
\end{equation}

%%%%%%%%%%%%%%%%%%%%%%%%%%%%%%%%%
%           CMD                 %
%%%%%%%%%%%%%%%%%%%%%%%%%%%%%%%%%
\noindent 
\textbf{Central Moment Discrepancy (CMD)}: Central Moment Discrepancy is another metric that measures the distance between source and target distributions. It not only considers the first moment and second moment, but also other higher-order moments. While MMD operates in a projected space, CMD operates in the representation space. If $P$ and $Q$ are two probability distributions and $X=\{X_1, X_2, ...., X_N\}$ and $Y=\{Y_1, Y_2, ...., Y_N\}$ are random vectors that are independent and identically distributed from $P$ and $Q$ and every component of the vector is bounded by $[a, b]$, CMD is then defined by:

\begin{equation}
    \begin{aligned}
    CMD(P, Q) = \frac{1}{|b-a|} \norm{E(X)-E(Y)}_2 \\ + \sum_{k=2}^{\infty} \frac{1}{|b-a|^k} \norm{c_k(X) - c_k(Y)}_2
    \end{aligned}
\end{equation}

where $E(X)$ is the expectation of X and $c_k$ is the $k-th$ order central moment, defined as:

\begin{equation}
    c_k(X) = E \bigg( \prod_{i=1}^{N} (X_i - E(X_i))^{r_i} \bigg)
\end{equation}

and $r_1 + r_2 + r_N = k$ and $r_1 .... r_N \geq 0$

% %%%%%%%%%%%%%%%%%%%%%%%%%%%%%%%%%%%%%%
% % Capture the other metrics here 
% % Makes it easier to reference in the text
% %%%%%%%%%%%%%%%%%%%%%%%%%%%%%%%%%%%%%%
\subsection{Other Measures}

\noindent 
\textbf{Bhattacharya Coefficient: }
If $P$ and $Q$ are probability distributions, then the Bhattacharya coefficient and Bhattacharya distance are defined as:

\begin{equation}
    Bhattacharya(P, Q) = \sum_{x} \sqrt{P(x) Q(x)}
\end{equation}

\begin{equation}
    D_{Bhattacharya} = -log(Bhattacharya(P, Q))
\end{equation}

\noindent
\textbf{Term Vocabulary Overlap (TVO)}: This measures the proportion of target vocabulary that is also present in the source vocabulary. If $V_S$ is the source domain vocabulary and $V_T$ is the target domain vocabulary, then the Term Vocabulary Overlap between the source domain ($D_S$) and the target domain ($D_T$) is given by:

\begin{equation}
    TVO(D_S, D_T) = \frac{|V_S \bigcap V_T|}{|V_T|}
\end{equation}

\noindent 
\textbf{Word Vector Variance}:
Different contexts in which a word is used in two different datasets can be used as an indication of the divergence between two datasets. Let $\vec{w}_{src}^{i}$ denote the  word embedding of word $i$ in source domain and $\vec{w}_{trg}^{i}$ is the word embedding of the same word in the target domain. Let $d$ be the dimension of the word embedding. The word vector variance between the source domain ($D_S$) and the target domain ($D_T$) is given by:

\begin{equation}
    WVV(D_S, D_T) = \frac{1}{|V_S| * d} \sum_{i}^{|V_s|} \norm{w_{src}^i - w_{trg}^i}_2^2
\end{equation}
\section{Model Hyperparameters}
\label{sec:hyperparams}
For POS, NER and Sentiment Analysis models, we do a grid search of learning rate in \{1e-01, 1e-05, 5e-05\} and dropout in $\{0.2, 0.3, 0.4, 0.5\}$ and number of epochs in $\{25, 50\}$. 
% MinCR: dangling sentence.  Attach it as a modifier to something else.
PAD requires a domain discriminator. We sample as many samples in the target domain as the source domain \cite{Ruder2017DataSS} and train a DistilBERT based classifier. For every domain discriminator we do a grid search of learning rate in \{{\it 1e-05, 5e-05}\}, dropout in $\{0.4, 0.5\}$ and number of epochs in $\{10,25\}$. For POS and NER, we monitor the macro F-Score; for domain discrimination, we monitor the accuracy scores. We chose the best model after the grid search for all subsequent calculations.
For training the models we use the Adam Optimiser \cite{kingma2014adam} with the $\beta_1=0.9$ and $\beta_2=0.99$  and $\epsilon$ as 1e-8. We use HuggingFace Transformers \cite{Wolf2019HuggingFacesTS} for all our experiments.
\section{Cross-Domain Performances}
\label{sec:cross-domain-perfs}

\subsection{Parts of speech tagging}
Table \ref{tab:pos-best-models-hyperparams} shows the hyper parameters for the best model for POS and Table \ref{tab:pos-cross-domain-perfs} shows the cross domain performances. 

\begin{table*}
    \small % Sets the font size to small on the entire table
    \centering
    \begin{tabular}{p{3cm} p{2cm} p{2cm} p{2cm} p{2cm}}
         \hline
         \textbf{Dataset} & \textbf{Epochs} & \textbf{Learning Rate} & \textbf{Dropout} & \textbf{Fscore} \\ \hline 
         EWT-answers & 50 & $5 \times 10^{-5}$ & 0.4 & 95.38 \\ 
         EWT-email & 50 & $1 \times 10^{-5}$ & 0.3 & 96.62 \\ 
         EWT-newsgroup & 50 & $5 \times 10^{-5}$ & 0.5 & 95.92 \\ 
         EWT-reviews & 50 & $5 \times 10^{-5}$  & 0.4  & 96.97 \\ 
         EWT-weblog & 50 & $5 \times 10^{-5}$ & 0.3 & 97.03 \\ 
         GUM & 50 & $1 \times 10^{-5}$ & 0.3 & 95.73\\ 
         LINES & 50 & $5 \times 10^{-5}$ & 0.3 & 97.38 \\ 
         PARTUT & 50  & $1 \times 10^{-5}$ & 0.4 & 97.06 \\ 
         \hline 
    \end{tabular}
    \caption{Model performance and hyper-parameters producing the best model for Parts of Speech Tagging trained using DistilBERT as the base model. The datasets are from the Universal Dependencies Corpus (UD) \cite{nivre-etal-2016-universal}. 5 corpora are from the English Word Tree (EWT) portion which are EWT-answers, EWT-email, EWT-newgroup, EWT-reviews, EWT-weblog.}.
    \label{tab:pos-best-models-hyperparams}
\end{table*}

\begin{table*}
    \small % Sets the font size to small on the entire table
    \centering
    \begin{tabular}{ c p{1.5cm} p{1cm} p{1.5cm} p{1cm} p{1cm} p{1cm} p{1cm} p{1.5cm} }
        \hline 
         \textbf{Source/Target} & EWT-answers & EWT-email & EWT-newsgroup & EWT-reviews & EWT-weblog & GUM & LINES & PARTUT \\
         \hline
         EWT-answers & 95.38 & 93.96 & 94.02 & 95.83 & 95.64 & 93.58 & 93.86 & 92.06 \\ 
         EWT-email & 94.11 & 96.62 & 94.40 & 95.42 & 95.37 & 93.08 & 93.98 & 93.47  \\ 
         EWT-newsgroup & 94.71 & 95.07 & 95.92 & 95.31 & 96.80 & 93.82 & 93.83 & 92.74  \\
         EWT-reviews & 94.99 & 94.51 & 94.56 & 96.97 & 95.55 & 93.07 & 94.27 & 92.62 \\ 
         EWT-weblog & 95.38 & 93.96 & 94.02 & 95.83 & 95.64 & 93.58 & 93.87 & 92.06  \\ 
         GUM & 91.63 & 92.59 & 91.75 & 93.55 & 93.56 & 95.73 & 93.54 & 93.12 \\ 
         LINES & 89.79 & 89.77 & 88.76 & 92.39 & 90.77 & 91.75 & 97.38 & 92.68   \\ 
         PARTUT & 89.27 & 89.54 & 89.56 & 91.28 & 92.27 & 90.65 & 92.97 & 96.65   \\ 
         \hline 
    \end{tabular}
    \caption{ Cross-domain performance for POS tagging. The best model for each source domain is tested on the test dataset of the same domain and all other domains.}
    \label{tab:pos-cross-domain-perfs}
\end{table*}

\subsection{Named Entity Recognition}
\label{sec:ner-cross-domain-perf}
Table \ref{tab:ner-best-models-hyperparams} shows the hyper parameters for the best model for NER and Table \ref{tab:ner-cross-domain-perfs} shows the cross domain performances.

\begin{table*}
    \small % Sets the font size to small on the entire table
    \centering
    \begin{tabular}{p{3cm} p{2cm} p{2cm} p{2cm} p{2cm}}
         \hline
         \textbf{Dataset} & \textbf{Epochs} & \textbf{Learning Rate} & \textbf{Dropout} & \textbf{Fscore} \\ \hline 
         CONLL-2003 & 50 & $5 \times 10^{-5}$ & 0.5 & 0.90  \\ 
         WNUT & 25 & $5 \times 10^{-5}$ & 0.5 & 0.50 \\ 
         Onto-BC & 50 & $5 \times 10^{-5}$ &  0.5 & 0.82 \\ 
         Onto-BN & 50 & $1 \times 10^{-5}$ & 0.3 & 0.89 \\ 
         Onto-MZ & 50 & $1 \times 10^{-5}$ & 0.3 & 0.86 \\ 
         Onto-NW & 25 & $5 \times 10^{-5}$ & 0.4 & 0.89 \\ 
         Onto-TC & 50 & $1 \times 10^{-5}$ & 0.5& 0.75 \\ 
         Onto-WB & 50 & $5 \times 10^{-5}$ & 0.4 & 0.63 \\ 
         \hline 
    \end{tabular}
    \caption{Model performance and hyper-parameters for Named Entity Recognition trained using DistilBERT as the base model. The datasets are CONLL-2003, Emerging and Rare Entity Recognition twitter dataset (WNUT), and six different sources of text in Ontonotes v5 \cite{hovy-etal-2006-ontonotes}}
    \label{tab:ner-best-models-hyperparams}
\end{table*}

\begin{table*}
    \small % Sets the font size to small on the entire table
    \centering
    \begin{tabular}{ c c c c c c c c c }
        \hline 
         \textbf{Source/Target} & CONLL & WNUT & ONTO-BC & ONTO-BN & ONTO-MZ & ONTO-NW & ONTO-TC & WB \\
         \hline
         CONLL 2003 & 0.90 & 0.37 & 0.54 & 0.65 & 0.59 & 0.54 & 0.51 & 0.41\\ 
         WNUT & 0.66 & 0.50 & 0.40 & 0.44 & 0.49 & 0.42 & 0.49 & 0.33 \\ 
         ONTO-BC  &  0.48	& 0.31	& 0.82	& 0.81	& 0.77	& 0.74	& 0.72	& 0.45 \\
         ONTO-BN  &  0.53	& 0.37	& 0.77	& 0.89	& 0.76	& 0.79	& 0.76	& 0.47  \\ 
         ONTO-MZ  &  0.49	& 0.29	& 0.72	& 0.78	& 0.86	& 0.75	& 0.69	& 0.45 \\ 
         ONTO-NW  &  0.52	& 0.32	& 0.73	& 0.86	& 0.73	& 0.89	& 0.76	& 0.46 \\ 
         ONTO-TC  &  0.51	& 0.37	& 0.61	& 0.64	& 0.57	& 0.55	& 0.75	& 0.41 \\ 
         ONTO-WB  &  0.43	& 0.12	& 0.52	& 0.63	& 0.54	& 0.57	& 0.52	& 0.63 \\ 
         \hline 
    \end{tabular}
    \caption{ Cross-domain performance for NER. The best model for each source domain is tested on the test dataset of the same domain and all other domains.}
    \label{tab:ner-cross-domain-perfs}
\end{table*}

\subsection{Sentiment Analysis}
\label{sec:sa-cross-domain-perf}
Table \ref{tab:sa-best-models-hyperparams} shows the hyper parameters for the best model for Sentiment analysis and Table \ref{tab:sa-cross-domain-perfs} shows the cross domain performances.

\begin{table*}
    \small % Sets the font size to small on the entire table
    \centering
    \begin{tabular}{p{3cm} p{2cm} p{2cm} p{2cm} p{2cm}}
         \hline
         \textbf{Dataset} & \textbf{Epochs} & \textbf{Learning Rate} & \textbf{Dropout} & \textbf{Fscore} \\ \hline 
         Apparel & 25 & $1 \times 10^{-5}$ &  0.4 & 91.25 \\ 
         Baby & 50 & $5 \times 10^{-5}$ & 0.4 & 93.75 \\ 
         Books & 50  & $1 \times 10^{-5}$  & 0.4 & 92  \\ 
         Camera/Photo & 25 & $1 \times 10^{-5}$ & 0.4 & 92 \\ 
         MR & 50 & $5 \times 10^{-5}$ & 0.3 & 82.5 \\ 
         \hline 
    \end{tabular}
    \caption{Model performance and hyper-parameters for Sentiment Analysis with DistilBERT as the base model. We chose 5 out of 16 datasets from \cite{liu-etal-2017-adversarial} which are Apparel, Baby, Books, Camera/Photo, and MR.}
    \label{tab:sa-best-models-hyperparams}
\end{table*}

\begin{table*}
    \small % Sets the font size to small on the entire table
    \centering
    \begin{tabular}{ c c c c c c}
        \hline 
         \textbf{Source/Target} & Apparel & Baby & Books & Camera/Photo & MR \\
         \hline
         Apparel & 0.91	& 0.9100    & 0.85    & 0.87    & 0.77\\ 
         Baby    & 0.89	& 0.9375	& 0.86	  & 0.89	& 0.75 \\ 
         Books   & 0.88	& 0.8875	& 0.92	  & 0.87	& 0.79   \\
         Camera/Photo  & 0.89	    & 0.89	  & 0.86	& 0.92	& 0.75 \\ 
         MR      & 0.76	& 0.76	    & 0.8375  & 0.74	& 0.83 \\ 
         \hline 
    \end{tabular}
    \caption{ Cross-domain performance for Sentiment Analysis. The best model for each source domain is tested on the test dataset of the same domain and all other domains.}
    \label{tab:sa-cross-domain-perfs}
\end{table*}
\section{Silhouette Scores and t-SNE Plots}

For calculating Silhouette scores  we use a subset of domain divergence measures that are metrics (a requirement of Silhouette scores) and can be calculated between single instances of text. We sample 200 points for each dataset as the time complexity increases exponentially with number of points.  We average the results over 5 runs. 

\begin{figure*}
    \centering
    \subfloat[POS \label{fig:pos-tsne}]{{\includegraphics[width=0.47\linewidth]{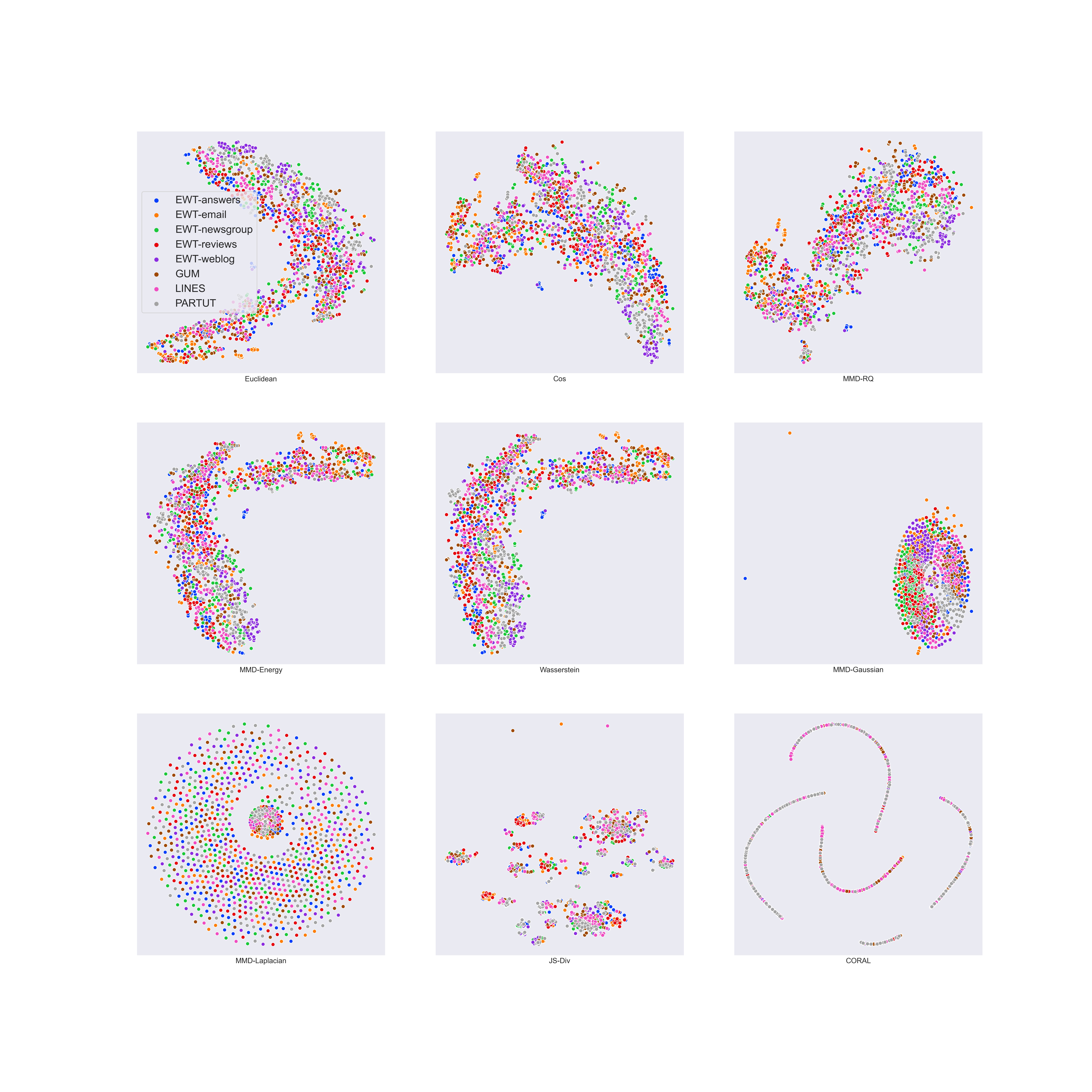} }}%
    \subfloat[NER \label{fig:ner-tsne}]{{\includegraphics[width=0.47\linewidth]{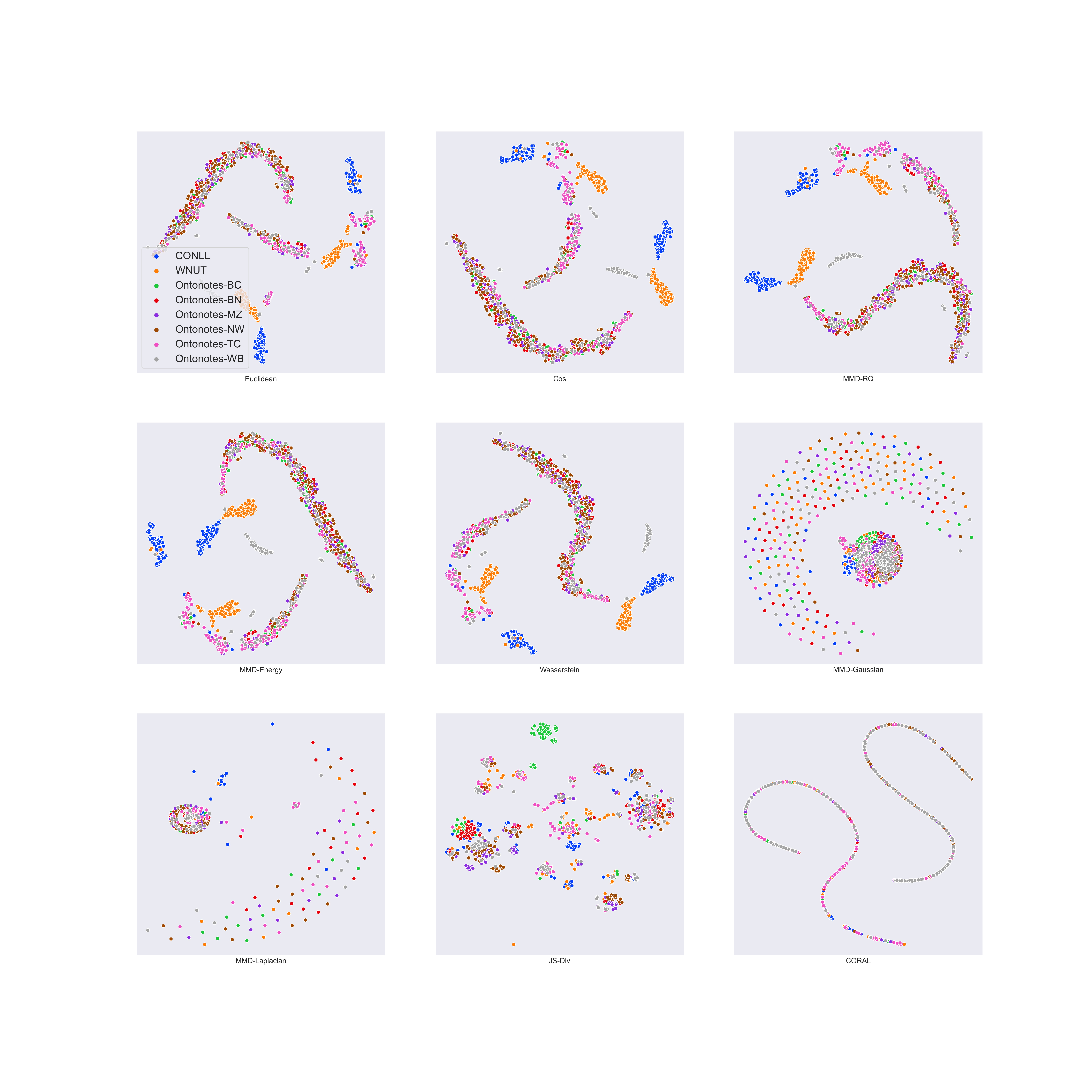} }}%

    \subfloat[SA \label{fig:sa-tsne}]{{\includegraphics[width=0.7\linewidth]{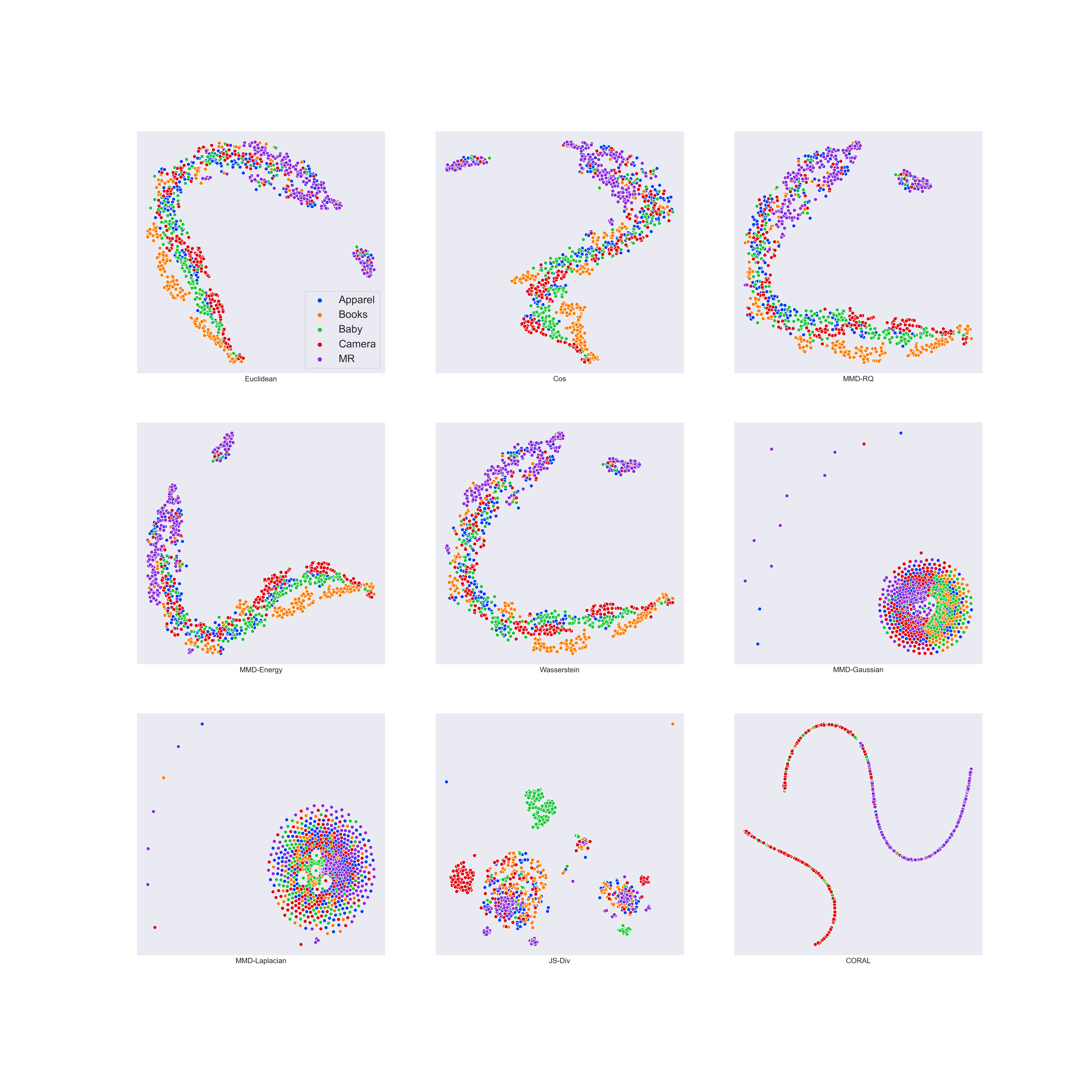} }}%
% MinCR: unhelpful caption.  Narrate the layout and colors in your caption.
    \caption{t-SNE plots for different tasks. }
\end{figure*}

We plot the t-SNE plots for POS (Figure \ref{fig:pos-tsne}), NER (Figure \ref{fig:ner-tsne}) and Sentiment Analysis (SA) (Figure \ref{fig:sa-tsne}). We sample 200 points from each of the datasets for the plot. Wherever relevant, we use DistilBERT \cite{Sanh2019DistilBERTAD} representations for calculations.
\label{sec:tsne-plots}

% \begin{figure*}
%     \centering
%     \includegraphics[width=\textwidth]{fig/pos-tsne.jpeg}
%     \caption{}
%     \label{fig:pos-tsne}
% \end{figure*}

% \begin{figure*}
%     \centering
%     \includegraphics[width=\textwidth]{fig/ner-tsne.jpeg}
%     \caption{t-SNE plots for NER corpora. }
%     \label{fig:ner-tsne}
% \end{figure*}

% \begin{figure*}
%     \centering
%     \includegraphics[width=\textwidth]{fig/sa-tsne.jpeg}
%     \caption{t-SNE plots for SA corpora. }
%     \label{fig:sa-tsne}
% \end{figure*}

\end{document}